\documentclass[5p,times]{elsarticle}
\usepackage{amsmath,amssymb,amsfonts}
\usepackage{subcaption}
\usepackage{booktabs}   
\usepackage{adjustbox}
\usepackage{tabularx}   
\usepackage{multicol}   
\usepackage{multirow}   
\usepackage{color}      
\usepackage{array}
\newcolumntype{C}{>{\centering\arraybackslash}X}
\usepackage{caption}
\usepackage{rotating}
\usepackage[ruled,vlined,linesnumbered]{algorithm2e}
\usepackage{url} 

\makeatletter
\AtBeginDocument{\DeclareMathVersion{bold}
\DeclareSymbolFont{NewLetters}{T1}{times}{m}{it}
\SetSymbolFont{operators}{bold}{T1}{times}{b}{n}
\SetSymbolFont{NewLetters}{bold}{T1}{times}{b}{it}
\SetMathAlphabet{\mathrm}{bold}{T1}{times}{b}{n}
\SetMathAlphabet{\mathit}{bold}{T1}{times}{b}{it}
\SetMathAlphabet{\mathbf}{bold}{T1}{times}{b}{n}
\SetMathAlphabet{\mathtt}{bold}{OT1}{pcr}{b}{n}
\SetSymbolFont{symbols}{bold}{OMS}{cmsy}{b}{n}
\renewcommand\boldmath{\@nomath\boldmath\mathversion{bold}}}
\makeatother

\def\BibTeX{{\rm B\kern-.05em{\sc i\kern-.025em b}\kern-.08em
    T\kern-.1667em\lower.7ex\hbox{E}\kern-.125emX}}

\journal{Journal}

\begin{document}
\begin{frontmatter}
\title{
A Source Domain is All You Need: Source-Only Cross-OS Transfer Learning for APT Anomaly Detection via Semantic Alignment and Optimal Transport}
\author[nyu]{Sidahmed Benabderrahmane\corref{cor1}}
\ead{sidahmed.benabderrahmane@nyu.edu}
\author[uqam]{Petko Valtchev}
\ead{valtchev.petko@uqam.ca}
\author[edin]{James Cheney}
\ead{jcheney@inf.ed.ac.uk}
\author[nyu]{Talal Rahwan}
\ead{talal.rahwan@nyu.edu}

\cortext[cor1]{Corresponding author}%
\address[nyu]{New York University, NYUAD, Division of Science, Computer Science Department.}
\address[uqam]{University of Quebec in Montreal, Computer Science Department, Montreal.}
\address[edin]{University of Edinburgh, School of Informatics, Edinburgh.}

%

%

\begin{abstract}   
Advanced Persistent Threats (APTs) are stealthy, multi-stage cyberattacks carried out by highly skilled adversaries aiming to infiltrate and maintain access to critical systems over extended periods. Detecting APTs remains challenging due to the rarity and scarcity of labeled attack traces, severe class imbalance, and the difficulty of synthetically generating realistic behaviors. These issues become even more pronounced in cross-operating system (cross-OS) scenarios, where labeled data in the target domain is unavailable, impeding generalization across platforms, and a detector trained on one source platform must be deployed on a different target platform without access to target-domain labels. We study this source-only cross-OS APT detection problem using system-level provenance traces and propose a transport-based framework for ranking anomalous target processes under zero target supervision. The proposed framework first abstracts process behavior into structured natural-language descriptions and embeds them into a shared semantic space using pretrained language models. It then constructs a source-domain normality reference and scores target processes through three complementary views: semantic deviation from source-normal prototypes, structural deviation captured by graph autoencoding, and geometric deviation measured through Optimal Transport (OT). The core methodological contribution is an OT-based barycentric anomaly score that projects target embeddings onto the source-normal manifold and quantifies residual transport mismatch. We further introduce entropy-weighted, angle-aware, and density-aware OT variants to capture complementary forms of cross-domain deviation. The semantic and structural components are used as auxiliary evidence channels, while the OT score provides the main geometry-aware transfer mechanism. We evaluate the framework on DARPA Transparent Computing provenance data spanning Linux, Windows, BSD, and Android, covering multiple APT scenarios and twelve cross-OS transfer pairs. The evaluation compares multiple language-model backbones, OT variants, scoring paths, fusion strategies, and transfer anomaly-detection baselines using both ROC-AUC and nDCG for ranking-oriented security triage. Results show that transport-based scoring provides strong cross-OS generalization and that combining semantic, structural, and geometric evidence improves robustness under severe class imbalance. The findings demonstrate that source-only provenance modeling, when coupled with semantic abstraction and OT-based anomaly scoring, can support practical cross-platform APT detection without target-domain supervision.\\
\end{abstract}
 \end{frontmatter}  
 
\section{Introduction}

\textit{"If you know the enemy and know yourself, you need not fear the result of a hundred battles."} — Sun Tzu, \textit{The Art of War}. In the digital battlefield of modern cybersecurity, knowing the enemy means detecting stealthy, multi-stage intrusions known as \textit{Advanced Persistent Threats (APTs)} \cite{dong2025cybersecurity,saha2025expert,wang2025dynamic}. These attacks are orchestrated by highly skilled adversaries who leverage customized malware, lateral movement, and long-term persistence to evade traditional defenses \cite{buchta2024advanced,sharma2023advanced}. 

APT detection remains particularly challenging because attack traces are rare, labeled examples are scarce, class imbalance is severe, and realistic malicious behaviors are difficult to synthesize \cite{che2024systematic,SALIM2023e17156}. These challenges become even more pronounced in heterogeneous computing environments, where Security Operation Centers (SOCs) monitor fleets spanning multiple operating systems such as Linux, Windows, BSD, and Android. In such settings, labeled data may be available for one source platform but absent for another target platform, making it difficult to deploy supervised detectors across operating systems \cite{ghoson2025review,longo2025data}.

This paper studies the problem of \emph{source-only cross-OS APT detection}: detecting anomalous processes in an unlabeled target operating system using only provenance traces and labels from a source operating system. This setting is practically important because collecting and labeling representative APT traces for every platform is costly, time-consuming, and often infeasible. At the same time, many adversarial tactics, techniques, and procedures (TTPs), such as privilege escalation, command-and-control, credential access, and lateral movement, are semantically related across operating systems even when their low-level artifacts differ. This suggests that cross-OS detection may be possible if platform-specific syntax can be abstracted into transferable behavioral representations.

Achieving this goal is difficult for three reasons. First, a \emph{semantic gap} exists between OS-specific behavioral artifacts; for example, \texttt{bash} in Linux and \texttt{cmd.exe} in Windows may express related operational intent through different commands and execution contexts. Second, a \emph{structural gap} arises because process interactions, execution graphs, and provenance neighborhoods differ across operating systems. Third, the target domain is unlabeled, which prevents conventional supervised adaptation or validation-based calibration. As a result, a cross-OS detector must transfer knowledge from the source domain while avoiding assumptions that require target labels.

To address these challenges, we propose a source-only cross-OS anomaly detection framework for system-level provenance traces. The framework abstracts process behavior into structured natural-language descriptions, embeds these descriptions into a shared semantic space using pretrained language encoders, constructs a source-domain normality reference, and ranks target processes according to complementary anomaly evidence. The central methodological component is an Optimal Transport (OT)-based barycentric anomaly score that projects target embeddings onto the source-normal manifold and measures residual transport mismatch. We further introduce entropy-weighted, angle-aware, and density-aware variants of this score to capture alignment uncertainty, directional drift, and low-density projection behavior.

Importantly, we distinguish between adapted components and new contributions. The language-model embeddings and Variational Graph Autoencoder (VGAE) modules build on established representation-learning techniques and are used as auxiliary semantic and structural evidence channels. The main contribution of this work lies in formalizing the source-only cross-OS APT detection setting, introducing OT-based barycentric anomaly scoring for provenance embeddings, and systematically evaluating this setting across multiple operating-system transfer pairs.

The contributions of this paper are as follows:

\begin{itemize}
    \item \textbf{Source-only cross-OS APT detection problem.}
    We define and formalize source-only cross-OS APT detection, where labeled source-domain provenance traces are available but the target operating system has no supervision. This setting reflects realistic security operations where representative attack labels are unavailable for every platform.

    \item \textbf{Transport-based barycentric anomaly scoring.}
    We introduce an OT-based anomaly score for cross-OS provenance embeddings. The score projects target processes onto the source-normal manifold through barycentric Optimal Transport and measures residual deviation. We further propose entropy-weighted, angle-aware, and density-aware variants to capture uncertainty, directional drift, and low-density projection behavior.

    \item \textbf{Multi-view source-only detection framework.}
    We design a framework that combines semantic, structural, and geometric evidence while respecting the zero-target-label constraint. The LLM-based semantic representation and VGAE-based structural modeling are used as adapted auxiliary components, whereas the OT-based score forms the central geometry-aware transfer mechanism.

    \item \textbf{Comprehensive cross-OS empirical benchmark.}
    We evaluate the framework on DARPA Transparent Computing provenance traces across four operating systems, two APT scenarios, and twelve source-target transfer pairs. The study reports ROC-AUC and nDCG, and compares multiple language-model backbones, OT variants, scoring paths, fusion strategies, and source-only anomaly-detection baselines.
\end{itemize}

\begin{figure}[t]
    \centering
    \includegraphics[width=1\linewidth]{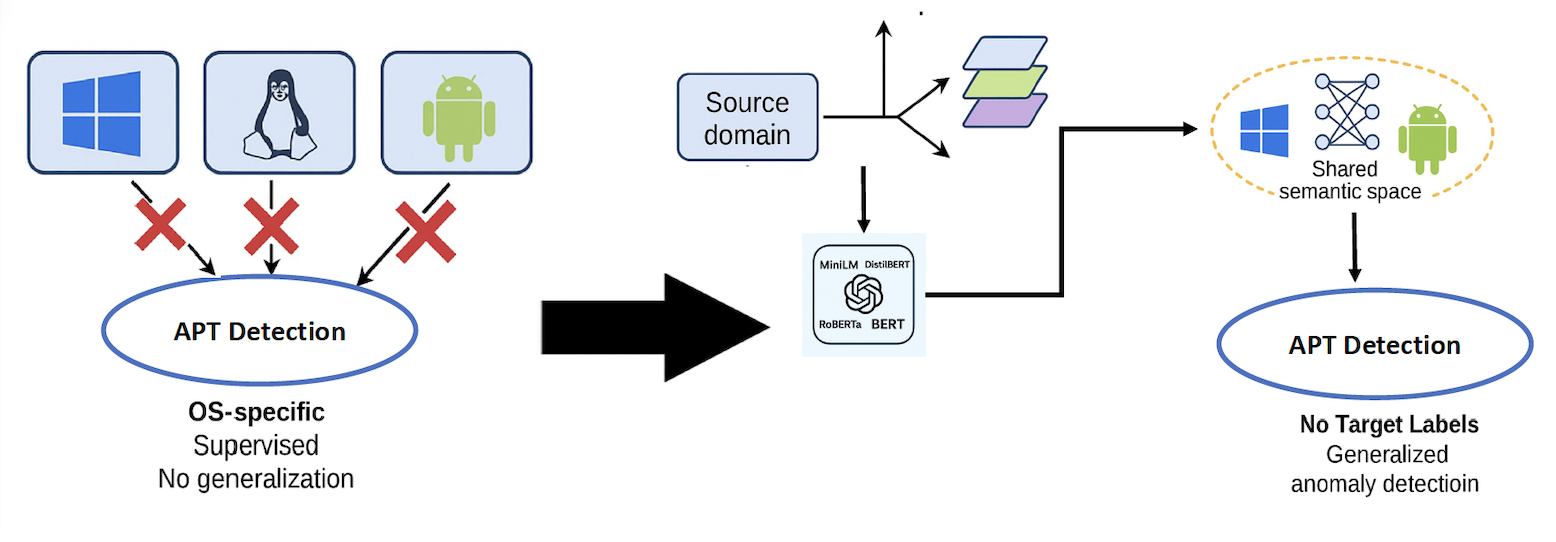}
    \caption{
    Conventional OS-specific APT detection pipelines require target-domain training or platform-specific supervision. The proposed framework addresses source-only cross-OS detection by combining semantic, structural, and OT-based geometric anomaly evidence under zero target supervision.}
    \label{fig:limitation}
\end{figure}

Figure~\ref{fig:limitation} motivates the proposed setting by contrasting conventional OS-specific APT detection pipelines with the source-only cross-OS setting considered in this work. Existing approaches are typically coupled to a specific operating system or require target-domain supervision, which limits their applicability in heterogeneous environments \cite{Flash24,Benabderrahmane2025RankingenhancedAD}. In contrast, our framework transfers source-domain provenance knowledge to an unlabeled target OS through semantic abstraction, structural modeling, and OT-based geometric scoring.

We evaluate the proposed framework on real-world cybersecurity datasets from the DARPA Transparent Computing (TC) program, spanning Linux, Windows, Android, and BSD operating systems. The results show that transport-based scoring provides strong cross-OS generalization and that combining semantic, structural, and geometric evidence improves robustness under severe class imbalance. The detected anomalous processes also show meaningful alignment with adversarial tactics from the MITRE ATT\&CK framework, supporting the operational relevance of the approach for cross-platform APT triage. 
\section{Related Work}

\subsection{APT and Provenance-Based Anomaly Detection}

Advanced Persistent Threats (APTs), including campaigns such as \textit{Stuxnet}, \textit{APT28}, and \textit{SolarWinds}, have demonstrated the risks of long-dwell, multi-stage intrusions against critical infrastructures and enterprise systems \cite{saha2025expert,singhal2025advanced,buchta2024advanced,sharma2023advanced}. Unlike short-lived attacks, APTs often involve stealthy sequences of actions such as initial compromise, privilege escalation, lateral movement, command-and-control communication, and persistence. This makes detection difficult because malicious activity may appear individually benign while becoming suspicious only when analyzed as part of a broader behavioral context.

Traditional anomaly detection methods, including statistical models, one-class classifiers, isolation forests, and reconstruction-based models, have been widely used in cybersecurity settings. However, APT detection remains challenging because attack instances are rare, labeled traces are scarce, class imbalance is extreme, and attack behaviors evolve over time \cite{che2024systematic,SALIM2023e17156,DBLP:journals/fgcs/BenabderrahmaneHVCR24,amine2024optimization}. Deep learning methods, including convolutional models, autoencoders, and attention-based architectures, have improved detection performance in several settings, but they often require large training corpora, careful tuning, and environment-specific feature engineering \cite{app12136816,abuodeh2021novel,9496635,satyapanich2020casie}. Unsupervised approaches reduce the need for attack labels, but they remain prone to high false positives and weak transferability when the deployment environment differs from the training environment \cite{Benabderrahmane21,segurola2024unsupervised}.

System-level provenance data has become an important substrate for APT detection because it captures dependencies among processes, files, sockets, and system activities. Provenance-based methods model causal or behavioral relationships and can therefore reveal suspicious multi-step behavior that may not be visible from isolated events. Recent systems have used graph learning and representation learning over provenance graphs to detect intrusions and anomalous behaviors \cite{Flash24,BerradaCBMMTW20,Benabderrahmane2025RankingenhancedAD}. However, most existing provenance-based detectors are designed for a fixed operating system or deployment environment. They typically assume that training and testing data share the same feature space, event vocabulary, or graph semantics. This limits their applicability to heterogeneous SOC environments where Windows, Linux, BSD, and Android endpoints must be monitored jointly.

\subsection{Cross-Domain and Transfer Learning in Cybersecurity}

Transfer learning and domain adaptation have been explored in cybersecurity to reduce dependence on labeled data in new environments. Existing studies have considered transfer across network traffic domains, malware families, intrusion datasets, and related security contexts \cite{BenabderrahmaneR25}. For example, adversarial domain adaptation has been used for network intrusion detection by pretraining on source traffic and adapting to target traffic \cite{singla2020transfer}. Feature-distance and deep transfer approaches have also been studied for identifying unknown attacks or transferring security models across related domains \cite{zhao2019cross}. In malware analysis, vision-based transfer learning has been used to extract features from binary images using pretrained convolutional networks \cite{krizhevsky2012imagenet}. Other approaches include cross-domain learning over assembly code, sequential attack modeling with hidden Markov models, and generative augmentation for low-frequency security events \cite{pei2020xda,chadza2020hidden,yang2022cyberdiffusion}.

Despite these advances, most transfer-learning approaches in security differ from the setting considered in this paper. First, many methods operate on network-level or malware-level data rather than system-level provenance traces. Second, several approaches assume access to target-domain samples for adaptation, calibration, or validation, and some require target labels. Third, cross-domain transfer is often studied between datasets or traffic distributions, rather than across operating systems where the event vocabulary, executable names, process semantics, and graph structure may differ substantially. In contrast, our setting is source-only and cross-OS: the detector must rank anomalous processes in an unlabeled target operating system using only source-domain provenance information.

\subsection{Semantic and Graph Representation Learning for Security Traces}

Representation learning has become central to security analytics because raw security events are sparse, heterogeneous, and difficult to compare directly across environments. Language models such as BERT, RoBERTa, DistilBERT, and MiniLM have shown that textual or sequence-based inputs can be embedded into dense semantic spaces that preserve contextual similarity \cite{devlin2019bert,liu2019roberta,sanh2019distilbert,wang2020minilm}. In cybersecurity, this motivates the use of structured behavioral descriptions to represent process activity, command execution, file access, and network communication in a form that can be compared semantically across platforms.

Graph representation learning provides a complementary view by modeling relations among entities or samples. Graph neural networks and graph autoencoders have been used to learn latent representations from relational data, including process graphs and provenance structures \cite{kipf2016semi,kipf2016vgae,Flash24}. Such methods are useful for detecting structural deviations, such as abnormal neighborhoods or unexpected process interactions. However, language-model embeddings and graph autoencoders are not, by themselves, sufficient to solve source-only cross-OS APT detection. Language models provide semantic abstraction, while graph models provide structural evidence, but neither directly defines a source-to-target anomaly score under zero target supervision. In our framework, these components are therefore used as adapted auxiliary evidence channels rather than as the main methodological novelty.

\subsection{Optimal Transport for Domain Alignment and Anomaly Scoring}

Optimal Transport (OT) provides a principled framework for comparing probability distributions by estimating the minimum cost required to move mass from one distribution to another. Entropic regularization and Sinkhorn-based computation have made OT scalable for machine learning applications \cite{cuturi2013sinkhorn,peyre2019computational}. OT has been used in domain adaptation and representation alignment because it can compare source and target distributions without requiring pointwise correspondences. This makes it particularly attractive for cross-domain anomaly detection, where the source and target distributions may be shifted but still share underlying semantic or geometric structure.

In this work, we use OT not only as a distribution-alignment tool but also as an anomaly-scoring mechanism. Target process embeddings are projected onto a source-normal manifold through barycentric transport, and their residual deviation from this projection is used as a geometric anomaly score. This differs from standard adaptation settings where OT is primarily used to align representations before training a classifier. Our formulation is source-only and label-free in the target domain: the transport map is used to quantify how far each target process deviates from source-normal behavior. We further extend this score with entropy-weighted, angle-aware, and density-aware variants to capture alignment uncertainty, directional drift, and low-density projection behavior.

\subsection{Positioning of This Work}

The above literature provides important building blocks for APT detection, transfer learning, semantic representation, graph modeling, and distribution alignment. However, existing work does not directly address the source-only cross-OS APT detection setting considered here, where labeled provenance data are available only for a source operating system and the target operating system has no supervision. Existing APT detectors are often \emph{within-domain}, transfer-learning methods often assume target adaptation or operate on network-level data, and representation-learning methods do not by themselves define a cross-OS anomaly score.

Our contribution is therefore not to claim novelty for language models or graph autoencoders individually. Instead, we formalize the source-only cross-OS APT detection problem and introduce OT-based barycentric anomaly scoring as a geometry-aware transfer mechanism for provenance embeddings. The resulting framework combines semantic, structural, and geometric evidence while preserving the zero-target-label constraint, and is evaluated across multiple operating systems, attack scenarios, embedding backbones, scoring paths, and transfer baselines.
\section{Problem Formulation}

We consider the problem of \emph{source-only cross-operating-system anomaly detection} for Advanced Persistent Threats (APTs). The objective is to rank anomalous processes in an unlabeled target operating system by using only provenance data and supervision available from a source operating system. This setting reflects realistic cybersecurity deployments where representative attack labels may exist for one platform but are unavailable for another platform.

Let
\[
\mathcal{D}_S = \{(x_i^S, y_i^S)\}_{i=1}^{n_S}
\]
denote the source domain, where each process instance \(x_i^S\) is collected from a source operating system and \(y_i^S \in \{0,1\}\) denotes its label, with \(0\) indicating benign behavior and \(1\) indicating anomalous or attack-related behavior. Let
\[
\mathcal{D}_T = \{x_j^T\}_{j=1}^{n_T}
\]
denote the target domain, consisting of unlabeled process instances collected from a different operating system. Target labels are assumed to be unavailable during training, scoring, model selection, and fusion; they are used only for offline evaluation.

Each process instance \(x\) is derived from system-level provenance telemetry, such as process events, executed routines, file operations, network flows, and process interactions. Depending on the scoring path, \(x\) may be represented as a structured behavioral sentence, a node in a process graph, or an embedding vector. We denote by \(\phi(\cdot)\) a representation function that maps process instances into a shared representation space:
\[
z = \phi(x), \qquad z \in \mathcal{Z}.
\]
The source and target embeddings are therefore denoted by
\[
\mathcal{Z}_S = \{\phi(x_i^S)\}_{i=1}^{n_S}, 
\qquad
\mathcal{Z}_T = \{\phi(x_j^T)\}_{j=1}^{n_T}.
\]

Because the target domain is unlabeled, the detector cannot rely on target-domain supervision or validation-based calibration. Instead, it must construct a source-domain reference of normal behavior. We define the source-normal subset as
\[
\mathcal{D}_S^{0} = \{x_i^S \in \mathcal{D}_S \mid y_i^S = 0\},
\]
with corresponding embeddings
\[
\mathcal{Z}_S^{0} = \{\phi(x_i^S) \mid y_i^S = 0\}.
\]
The goal is to learn an anomaly scoring function
\[
\mathcal{A}: \mathcal{Z}_T \rightarrow \mathbb{R}
\]
using only \(\mathcal{D}_S\), and in particular the source-normal reference \(\mathcal{D}_S^{0}\), such that higher scores indicate a greater likelihood that a target process is anomalous.

This setting differs from conventional supervised, semi-supervised, and domain-adaptation formulations. In supervised detection, labeled target-domain examples are available. In standard domain adaptation, unlabeled or partially labeled target data may be used for adaptation and hyperparameter selection. In contrast, our setting is \emph{source-only}: target samples may be embedded and scored at inference time, but their labels are never used to train, calibrate, or tune the detector.

The main difficulty comes from cross-OS distribution shift. Source and target processes may differ in three complementary ways. First, they exhibit \emph{semantic heterogeneity}, because operating systems use different commands, routines, and event vocabularies to express related behavior; for example, \texttt{bash} in Linux and \texttt{cmd.exe} in Windows may serve related execution roles. Second, they exhibit \emph{structural disparity}, because process graphs, execution neighborhoods, and provenance dependencies differ across platforms. Third, they exhibit \emph{geometric shift}, because even semantically related behaviors may occupy different regions of the embedding space.

We instantiate the anomaly scoring function through three complementary evidence channels. The semantic channel compares target process descriptions against source-normal behavioral prototypes. The structural channel measures deviations in graph-based process neighborhoods. The geometric channel uses Optimal Transport to project target embeddings onto the source-normal manifold and quantify residual transport mismatch. The final score is obtained by fusing the three path-level scores:
\begin{equation}
\mathcal{A}(x_j^T) =
\operatorname{Fuse}
\left(
\mathcal{A}_{\mathrm{sem}}(x_j^T),
\mathcal{A}_{\mathrm{str}}(x_j^T),
\mathcal{A}_{\mathrm{OT}}(x_j^T)
\right).
\end{equation}

In the main framework, we use a calibration-free max-fusion rule:
\begin{equation}
\mathcal{A}(x_j^T) =
\max
\left\{
\mathcal{A}_{\mathrm{sem}}(x_j^T),
\mathcal{A}_{\mathrm{str}}(x_j^T),
\mathcal{A}_{\mathrm{OT}}(x_j^T)
\right\}.
\end{equation}
This choice is motivated by the zero-target-label setting, where learned or validation-tuned fusion is not reliable. Max-fusion acts as a conservative security rule: a process is prioritized if any complementary view provides strong evidence of abnormality. Alternative fusion strategies are evaluated empirically in the experimental section.
\section{Methodology}
\label{sec:methodology}

\subsection{Overview}
\label{subsec:method_overview}

We propose a source-only cross-OS APT detection framework for ranking anomalous target-domain processes without using target labels. Given a labeled source domain \(\mathcal{D}_S\) and an unlabeled target domain \(\mathcal{D}_T\), the framework constructs a source-normal behavioral reference and assigns each target process an anomaly score through three complementary evidence channels: semantic, structural, and geometric. The semantic channel measures deviation from source-normal behavioral prototypes in a language-model embedding space. The structural channel measures deviations from expected graph neighborhoods using a Variational Graph Autoencoder (VGAE). The geometric channel, which is the core methodological component of the framework, uses barycentric Optimal Transport (OT) to project target embeddings onto the source-normal manifold and score residual transport mismatch.

\begin{figure*}[t]
    \centering
    \includegraphics[width=0.7\linewidth]{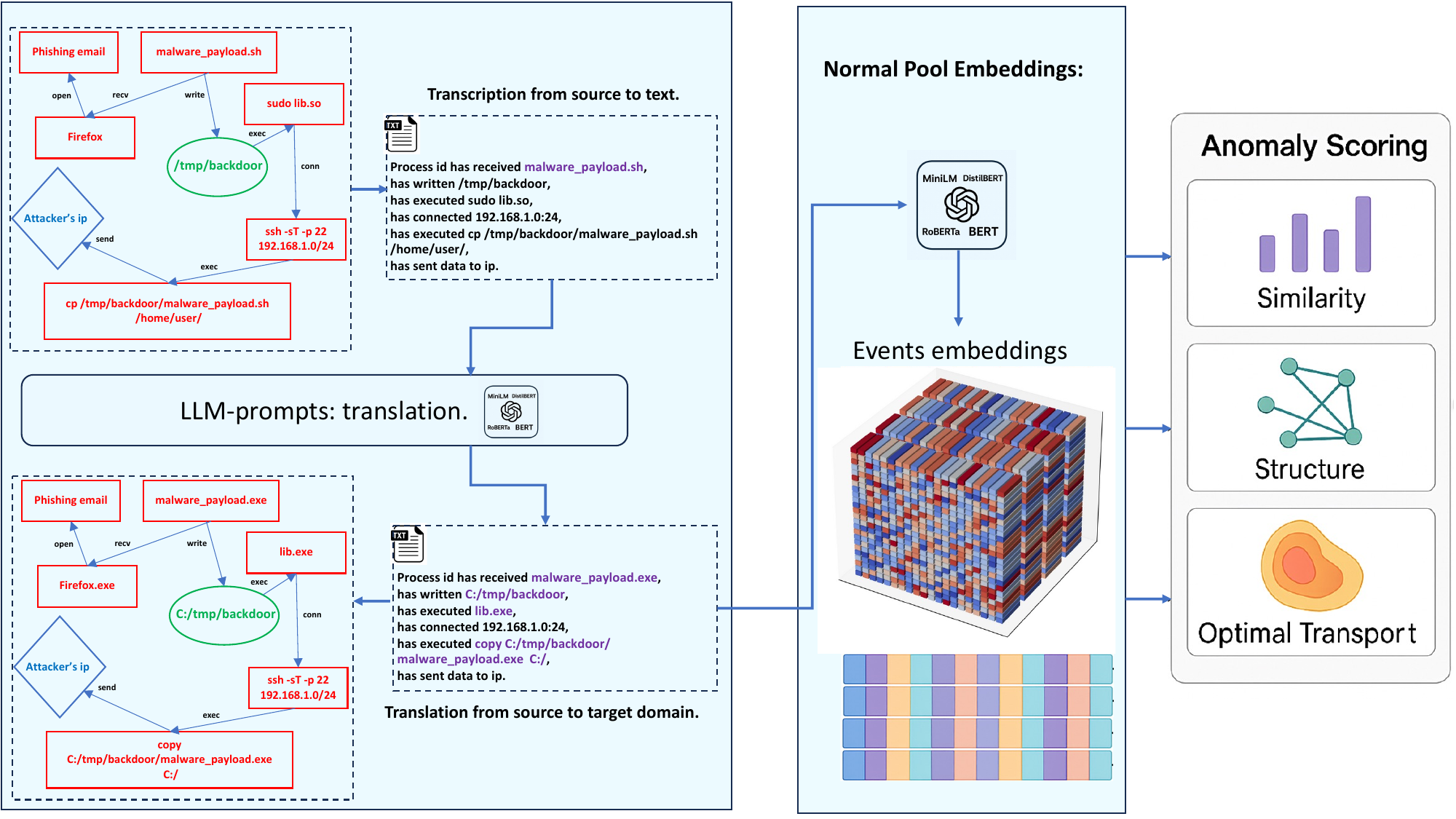}
    \caption{
    Source-only cross-OS transfer pipeline. Source-domain provenance traces are converted into structured behavioral descriptions, optionally translated into target-domain terminology, and embedded into a shared semantic space. A source-normal prototype memory is constructed from benign source processes and their target-aligned translated counterparts. Target processes are embedded and scored through semantic similarity, structural reconstruction, and OT-based barycentric projection without using target labels.}
    \label{fig:semantic-transfer}
\end{figure*}

The framework respects the source-only constraint throughout the pipeline. Source labels are used only to identify source-normal processes and build the reference memory. Target labels are never used for representation learning, score computation, score normalization, fusion, model selection, or hyperparameter tuning; they are used only for offline evaluation. Figure~\ref{fig:semantic-transfer} summarizes the overall cross-OS transfer pipeline.

Let \(\mathcal{D}_S^0 = \{x_i^S \mid y_i^S = 0\}\) denote the source-normal subset. For each language-model backbone \(l\), each process is encoded as
\[
z_i^{S,l} = g_l(s(x_i^S)), 
\qquad
z_j^{T,l} = g_l(s(x_j^T)),
\]
where \(s(\cdot)\) denotes a structured behavioral sentence and \(g_l(\cdot)\) is a pretrained sentence encoder. Source-normal prototypes are stored in a memory \(\mathcal{M}_S^{0,l}\), which contains original source-domain embeddings and, when translation is used, target-aligned translated embeddings. Target embeddings are denoted by
\[
\mathcal{Z}_T^l = \{z_j^{T,l} \mid x_j^T \in \mathcal{D}_T\}.
\]
For readability, we omit the superscript \(l\) when the context is clear.

\subsection{Behavioral Sentence Construction and Semantic Alignment}
\label{subsec:semantic_alignment}

Each process is represented from system-level provenance telemetry. A process instance \(x\) may include executed routines, system events, file operations, network flows, and parent-child process relations. We serialize these multi-view signals into a structured natural-language behavioral sentence:
\[
s(x) = \operatorname{Serialize}(e_1,e_2,\ldots,e_m),
\]
where each \(e_r\) denotes an observed provenance element. For example, a process may be represented as:

\begin{quote}
\small
\texttt{Process 123 performed event\_open and event\_write, executed /bin/bash and nmap, and sent data to 192.168.1.100:22.}
\end{quote}

This textual abstraction serves two purposes. First, it produces a common representation format for heterogeneous provenance views. Second, it reduces dependence on OS-specific feature names by exposing the behavioral intent of process activity.

To reduce cross-OS vocabulary mismatch, source-domain technical tokens may be translated into target-domain terminology using restricted prompt-based mapping. The prompt is constrained to translate discrete technical identifiers and does not ask the language model to generate new behaviors. This reduces hallucination risk because the model is used only for canonicalizing OS-specific vocabulary, not for inventing process activity. The resulting target-aligned source sentence is denoted by \(\tilde{s}(x_i^S)\).

For example, a Linux source sentence such as
\begin{quote}
\small
\texttt{Process 42 executed /bin/bash},\\ 
\texttt{loaded /usr/lib/lib.so},\\
\texttt{copied /tmp/backdoor/malware\_payload.sh}\\
\texttt{to /home/user/},\\
\texttt{opened /etc/passwd},\\
\texttt{and sent data through ssh to 192.168.1.10:22.}
\end{quote}
may be translated into a Windows-aligned sentence:
\begin{quote}
\small
\texttt{Process 42 executed cmd.exe,}\\
\texttt{loaded C:/Windows/System32/lib.exe,}\\
\texttt{copied C:/tmp/backdoor/malware\_payload.exe}\\
\texttt{to C:/Users/user/,}\\
\texttt{accessed an account-information file,}\\
\texttt{and sent data through an SSH client to 192.168.1.10:22.}
\end{quote}
The goal is not to alter the observed behavior, but to express source-domain actions using target-domain terminology when an equivalent concept exists. For instance, \texttt{/bin/bash} may be mapped to \texttt{cmd.exe}, a Linux shared object such as \texttt{lib.so} may be mapped to a Windows executable/library-style artifact such as \texttt{lib.exe}, and a copy operation such as \texttt{cp} may be expressed as \texttt{copy}. When no reliable equivalent exists, the original token is retained or mapped to a generic behavioral description.

For each embedding backbone \(l\), we construct a source-normal prototype memory:
\[
\mathcal{M}_S^{0,l} =
\left\{
g_l(s(x_i^S)),\,
g_l(\tilde{s}(x_i^S))
\;\middle|\;
x_i^S \in \mathcal{D}_S^0
\right\}.
\]
Here, \(g_l(s(x_i^S))\) denotes the embedding of the original source-domain behavioral sentence, while \(g_l(\tilde{s}(x_i^S))\) denotes the embedding of its target-aligned translated version. The memory \(\mathcal{M}_S^{0,l}\) therefore contains source-normal behavioral prototypes expressed both in the original source vocabulary and in the target-aligned vocabulary.

If translation is not used for a particular experiment, the memory reduces to the original source-normal embeddings. Duplicate prototypes can be removed or retained depending on implementation; in our experiments, the prototype memory is treated as a multiset because repeated benign behaviors reflect source-domain normality frequency.

\subsection{Semantic Similarity Scoring}
\label{subsec:semantic_score}

The semantic score measures how far a target process is from the closest source-normal behavioral prototype. For a target process \(x_j^T\), we compute
\[
z_j^T = g_l(s(x_j^T)).
\]
The semantic anomaly score is defined as the minimum cosine distance to the source-normal prototype memory:
\begin{equation}
\mathcal{A}_{\mathrm{sem}}^{(l)}(x_j^T)
=
\min_{z \in \mathcal{M}_S^{0,l}}
\left(
1 - \cos(z_j^T,z)
\right).
\label{eq:semantic_score}
\end{equation}
A low score indicates that the target behavior is semantically close to at least one source-normal behavior, whereas a high score indicates semantic deviation. This score is label-free with respect to the target domain and provides the first evidence channel of the framework.

\subsection{Structural Modeling via Source-Normal VGAE}
\label{subsec:vgae_score}

Semantic similarity captures pointwise behavioral deviation but does not explicitly model the local organization of processes in the embedding space. A target process may be suspicious not only because it is far from a source-normal prototype, but also because its neighborhood structure is inconsistent with normal source-domain geometry. To capture this effect, we use a Variational Graph Autoencoder (VGAE) as a structural evidence channel.

For each embedding backbone \(l\), we use the previously constructed source-normal prototype memory \(\mathcal{M}_S^{0,l}\). Thus, the structural graph is built over source-normal prototypes rather than over source processes alone. Let
\[
N_S^{0,l}=|\mathcal{M}_S^{0,l}|
\]
denote the number of source-normal prototypes for encoder \(l\). Depending on whether duplicate prototypes are removed, \(N_S^{0,l}\) can be smaller than or equal to \(2|\mathcal{D}_S^0|\).

We then construct a source-normal \(k\)-nearest-neighbor graph
\[
G_S^{0,l} = (\mathcal{V}_S^{0,l},\mathcal{E}_S^{0,l}),
\]
where each node corresponds to one prototype embedding in \(\mathcal{M}_S^{0,l}\). Edges are created between nearest neighbors under cosine similarity and are symmetrized to obtain an undirected graph. The neighborhood size is set as
\[
k = \max(k_{\min}, \lceil \rho N_S^{0,l} \rceil),
\]
where \(\rho\) controls graph density and \(k_{\min}\) avoids degenerate graphs for small source-normal memories.

Let
\[
\mathbf{X}_S^{0,l}\in\mathbb{R}^{N_S^{0,l}\times d_l}
\]
denote the node feature matrix of the source-normal prototype graph, where each row is the \(d_l\)-dimensional embedding of one source-normal prototype under encoder \(l\). Let
\[
\mathbf{A}_S^{0,l}\in\{0,1\}^{N_S^{0,l}\times N_S^{0,l}}
\]
denote the adjacency matrix of the source-normal \(k\)-nearest-neighbor graph.

The VGAE encoder maps node features and graph structure to a latent Gaussian distribution:
\begin{equation}
q_{\theta}(\mathbf{h}_i \mid \mathbf{X}_S^{0,l},\mathbf{A}_S^{0,l})
=
\mathcal{N}(\boldsymbol{\mu}_i,\operatorname{diag}(\boldsymbol{\sigma}_i^2)),
\qquad
\mathbf{h}_i \in \mathbb{R}^{r},
\end{equation}
where \(r\) is the VGAE latent dimension. The decoder reconstructs graph edges through an inner product:
\begin{equation}
\hat{A}_{ij}
=
\sigma(\mathbf{h}_i^\top \mathbf{h}_j),
\end{equation}
where \(\sigma(\cdot)\) is the sigmoid function.

The reconstruction loss is a binary cross-entropy over observed source-normal prototype edges and sampled non-edges:
\begin{equation}
\mathcal{L}_{\mathrm{recon}}
=
-
\sum_{(i,j)\in \mathcal{E}_S^{0,l}}
\log \hat{A}_{ij}
-
\alpha
\sum_{(i,j)\in \bar{\mathcal{E}}_S^{0,l}}
\log(1-\hat{A}_{ij}),
\label{eq:vgae_recon}
\end{equation}
where \(\bar{\mathcal{E}}_S^{0,l}\) denotes a sampled set of non-edge pairs from the source-normal prototype graph, and \(\alpha\) balances positive and negative reconstruction terms.

The complete VGAE objective is
\begin{equation}
\mathcal{L}_{\mathrm{VGAE}}
=
\mathcal{L}_{\mathrm{recon}}
+
\frac{1}{N_S^{0,l}}
D_{\mathrm{KL}}
\left(
q_{\theta}(\mathbf{H}\mid \mathbf{X}_S^{0,l},\mathbf{A}_S^{0,l})
\,\|\, 
\mathcal{N}(\mathbf{0},\mathbf{I})
\right),
\label{eq:vgae_loss}
\end{equation}
where \(\mathbf{H}=\{\mathbf{h}_i\}_{i=1}^{N_S^{0,l}}\) denotes the set of latent structural embeddings. The KL term regularizes the latent space toward a standard Gaussian prior.

At inference, each target process \(x_j^T\) is embedded using the same encoder \(g_l\) and inserted into the source-normal prototype graph by connecting it to its \(k\) nearest source-normal prototypes. This produces an augmented graph used only for scoring and does not use target labels. Let \(\mathcal{N}_S^{l}(j)\) denote the set of source-normal prototype neighbors of target process \(x_j^T\). The structural anomaly score is defined as the average reconstruction mismatch over this neighborhood:
\begin{equation}
\mathcal{A}_{\mathrm{str}}^{(l)}(x_j^T)
=
\frac{1}{|\mathcal{N}_S^{l}(j)|}
\sum_{i \in \mathcal{N}_S^{l}(j)}
\left|
A_{ji} - \hat{A}_{ji}
\right|,
\label{eq:structural_score}
\end{equation}
where \(A_{ji}=1\) for the inserted target--prototype edges and \(\hat{A}_{ji}\) is the VGAE reconstruction probability. A high reconstruction mismatch indicates that the VGAE, trained only on source-normal prototype structure, cannot represent the target process neighborhood as normal.

\subsection{Barycentric Optimal Transport for Geometric Scoring}
\label{subsec:ot_score}

The core methodological component of our framework is an Optimal Transport (OT)-based geometric anomaly score. While semantic similarity compares a target process to individual source-normal prototypes, OT compares the target embedding distribution to the source-normal embedding distribution and provides a soft geometric projection of each target point onto the source-normal support.

Optimal Transport provides a principled way to compare probability distributions by estimating a minimum-cost transport plan between them \cite{peyre2019computational}. Given two probability measures \(\mu\) and \(\nu\), the original Monge formulation seeks a deterministic map \(T\) that pushes \(\mu\) onto \(\nu\) while minimizing transport cost. Since such a map may not exist or may be difficult to compute, the Kantorovich relaxation instead seeks a transport plan \(\gamma \in \Pi(\mu,\nu)\), i.e., a joint distribution whose marginals are \(\mu\) and \(\nu\):
\begin{equation}
\gamma^\star
=
\arg\min_{\gamma \in \Pi(\mu,\nu)}
\int_{\mathcal{X}\times\mathcal{Y}} c(x,y)\, d\gamma(x,y),
\label{eq:kantorovich}
\end{equation}
where \(c(x,y)\) is a ground cost, typically based on Euclidean distance. In our setting, OT is not used to train a target-domain classifier. Instead, it is used as a source-only anomaly-scoring mechanism: target embeddings are softly projected onto the source-normal embedding support, and their residual deviation from this projection is used as a geometric anomaly score.

\begin{figure}[t]
    \centering
    \includegraphics[width=1\linewidth]{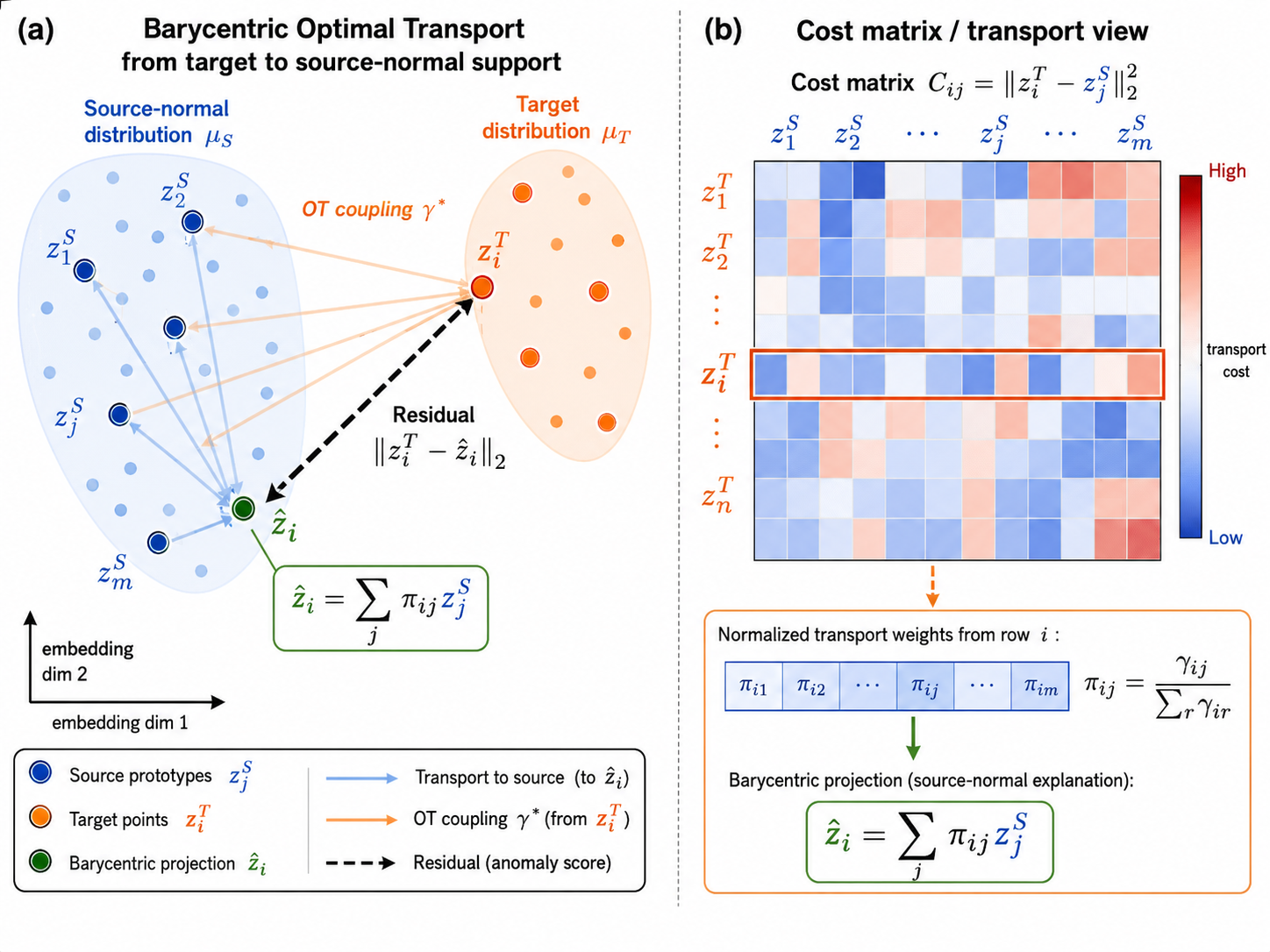}
    \caption{
    Illustration of the proposed OT-based anomaly score. 
    Left: a target embedding \(z_i^T\) is softly transported to the source-normal support through an OT coupling \(\gamma^\star\). Row-normalized transport weights \(\pi_{ij}\) define a barycentric projection \(\hat{z}_i\), interpreted as the source-normal explanation of the target point. The residual \(\|z_i^T-\hat{z}_i\|_2\) is used as a geometric anomaly score. 
    Right: the cost matrix \(C_{ij}=\|z_i^T-z_j^S\|_2^2\) and the row corresponding to \(z_i^T\) determine the normalized transport weights used in the barycentric projection.}
    \label{fig:ot_barycentric}
\end{figure}

Figure~\ref{fig:ot_barycentric} illustrates the intuition behind the proposed OT-based anomaly score. The left panel shows how a target embedding \(z_i^T\) is softly matched to multiple source-normal prototypes through the OT coupling \(\gamma^\star\). Rather than assigning the target point to a single nearest neighbor, OT produces a distribution of transport weights over the source-normal support. These weights define a barycentric projection \(\hat{z}_i\), which can be interpreted as the best source-normal explanation of the target process under the learned transport plan. The residual distance \(\|z_i^T-\hat{z}_i\|_2\) then measures how much of the target behavior remains unexplained by source-normal behavior.

The right panel of Figure~\ref{fig:ot_barycentric} shows the corresponding cost matrix \(C\), where each entry \(C_{ij}=\|z_i^T-z_j^S\|_2^2\) measures the transport cost between a target process and a source-normal prototype. Low-cost entries indicate source-normal prototypes that can plausibly explain a target process, whereas high-cost entries indicate poor matches. The OT coupling uses this cost structure to determine how target mass should be assigned to source-normal prototypes. In our anomaly detection setting, target processes that require high-cost or diffuse transport, or that remain far from their barycentric projection, are considered more suspicious.

For each embedding backbone \(l\), let
\[
\mathcal{Z}_T^{l} = \{z_i^{T,l}\}_{i=1}^{n}
\]
denote the target embeddings, and let
\[
\mathcal{M}_S^{0,l} = \{z_j^{S,l}\}_{j=1}^{m_l}
\]
denote the previously constructed source-normal prototype memory. This memory contains both original source-normal embeddings and their target-aligned translated counterparts. For readability, we omit the superscript \(l\) in the remainder of this subsection and write \(m=m_l\).

We define empirical measures over the target and source-normal embeddings:
\[
\mu_T = \sum_{i=1}^{n} a_i \delta_{z_i^T},
\qquad
\mu_S = \sum_{j=1}^{m} b_j \delta_{z_j^S},
\]
where \(\delta_z\) denotes the Dirac measure centered at embedding \(z\), i.e., a unit probability mass located at \(z\). Thus, \(\mu_T\) and \(\mu_S\) are empirical distributions over target and source-normal embeddings, respectively. Unless otherwise stated, we use uniform weights \(a_i=1/n\) and \(b_j=1/m\).

The pairwise transport cost between target and source-normal embeddings is defined as
\begin{equation}
C_{ij}
=
\|z_i^T - z_j^S\|_2^2.
\label{eq:cost_matrix}
\end{equation}
To obtain a smooth and computationally efficient coupling, we use entropic OT:
\begin{equation}
\gamma^\star
=
\arg\min_{\gamma \in \Pi(\mu_T,\mu_S)}
\langle \gamma, C \rangle
-
\varepsilon H(\gamma),
\label{eq:entropic_ot}
\end{equation}
where
\[
H(\gamma)
=
-\sum_{i,j}\gamma_{ij}\log(\gamma_{ij})
\]
is the entropy of the coupling, \(\varepsilon>0\) is the entropic regularization strength, and \(\Pi(\mu_T,\mu_S)\) denotes the set of couplings with marginals \(\mu_T\) and \(\mu_S\). Small values of \(\varepsilon\) yield sharper couplings, while larger values produce smoother transport plans.

Given the optimal coupling \(\gamma^\star\), we define row-normalized transport weights:
\begin{equation}
\pi_{ij}
=
\frac{\gamma_{ij}^{\star}}
{\sum_{r=1}^{m}\gamma_{ir}^{\star}},
\qquad
\sum_{j=1}^{m}\pi_{ij}=1.
\label{eq:conditional_weights}
\end{equation}
The normalization is important because \(\gamma^\star\) is a transport plan whose rows sum to the target marginal \(a_i\). The normalized weights \(\pi_{ij}\) therefore define the conditional source-normal distribution associated with target point \(z_i^T\).

The barycentric projection of \(z_i^T\) onto the source-normal support is then
\begin{equation}
\hat{z}_i
=
\sum_{j=1}^{m}
\pi_{ij} z_j^S.
\label{eq:barycentric_projection}
\end{equation}
This projection is a convex combination of source-normal prototypes and can be interpreted as the source-normal explanation of the target process under the learned transport plan.

The baseline OT anomaly score is the barycentric residual:
\begin{equation}
\mathcal{A}_{\mathrm{OT,res}}^{(l)}(x_i^T)
=
\|z_i^T - \hat{z}_i\|_2.
\label{eq:ot_residual}
\end{equation}
A large residual indicates that the target process cannot be well represented by a transported combination of source-normal behaviors, suggesting that it lies away from the source-normal manifold.

\subsection{Why the Barycentric Residual is a Principled Anomaly Score}
\label{subsec:ot_principle}

The barycentric OT residual can be interpreted as a source-normal manifold deviation score. Since \(\hat{z}_i\) is a convex combination of source-normal prototypes, it always lies in the convex hull of the source-normal support:
\[
\hat{z}_i \in \operatorname{conv}(\mathcal{M}_S^{0,l}).
\]
Therefore, the residual \(\|z_i^T-\hat{z}_i\|_2\) measures how far the target point remains from a source-normal explanation after OT-based soft alignment.

\paragraph{Geometric observation.}
For any target embedding \(z_i^T\) and any barycentric projection \(\hat{z}_i \in \operatorname{conv}(\mathcal{M}_S^{0,l})\),
\[
\|z_i^T-\hat{z}_i\|_2
\geq
\operatorname{dist}
\left(
z_i^T,
\operatorname{conv}(\mathcal{M}_S^{0,l})
\right).
\]
Thus, if a target point lies far from the source-normal convex support, every barycentric source-normal projection must incur a large residual. Conversely, if a normal target behavior is transferable across operating systems, semantic abstraction should place it close to the transported source-normal support, allowing a small barycentric residual.

This observation motivates the use of the OT residual as an anomaly score. Normal target processes are expected to admit a compact source-normal explanation after semantic alignment, while anomalous processes are expected to remain poorly explained by the transported source-normal manifold.

\subsection{OT Score Variants}
\label{subsec:ot_variants}

The baseline residual captures geometric mismatch, but different types of cross-OS deviation may appear as uncertain alignment, directional semantic drift, or projection into sparse source-normal regions. We therefore define three complementary OT variants.

\begin{figure}[t]
    \centering
    \includegraphics[width=1\linewidth]{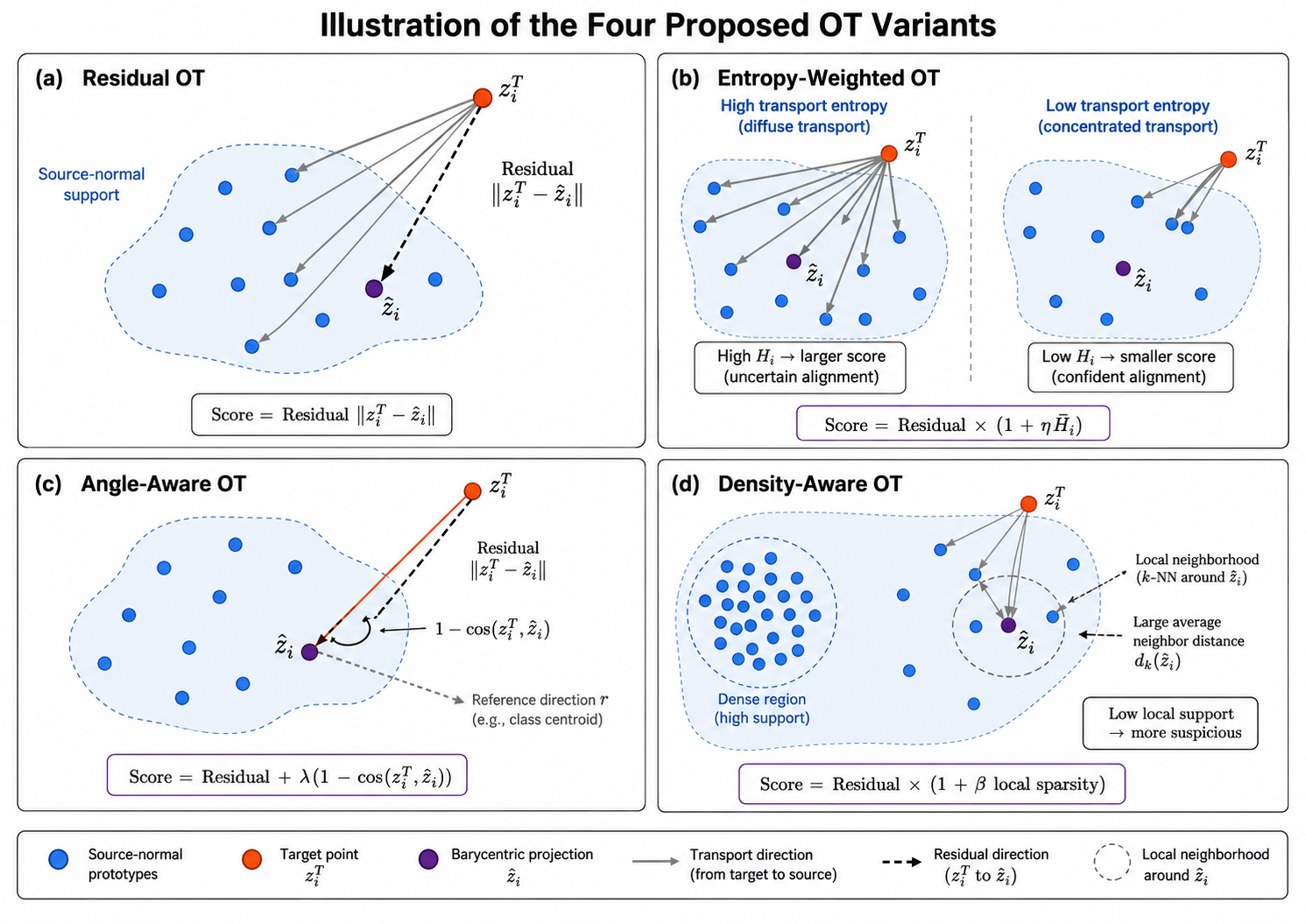}
    \caption{
    Illustration of the four proposed OT-based anomaly scoring variants. 
    (a) Residual OT measures the Euclidean distance between a target embedding \(z_i^T\) and its barycentric source-normal projection \(\hat{z}_i\). 
    (b) Entropy-weighted OT amplifies the residual when the row-normalized transport weights are diffuse, indicating uncertain alignment to many source-normal prototypes. 
    (c) Angle-aware OT combines the residual with an angular deviation term \(1-\cos(z_i^T,\hat{z}_i)\), capturing directional semantic drift. 
    (d) Density-aware OT increases the score when the barycentric projection lies in a sparse source-normal region, measured through the local \(k\)-NN neighborhood around \(\hat{z}_i\).}
    \label{fig:ot_variants}
\end{figure}

Figure~\ref{fig:ot_variants} illustrates the intuition behind the four OT-based scoring variants. All variants start from the same barycentric projection idea: a target process \(z_i^T\) is softly matched to the source-normal support and projected to \(\hat{z}_i\). Residual OT scores the Euclidean mismatch between the target point and this projection. Entropy-weighted OT increases the score when the transport weights are diffuse, indicating uncertain alignment. Angle-aware OT additionally penalizes directional semantic drift between \(z_i^T\) and \(\hat{z}_i\). Density-aware OT increases the score when the projection falls in a sparse region of the source-normal support.

These variants are complementary: the residual term captures geometric distance, the entropy term captures alignment uncertainty, the angular term captures directional mismatch, and the density term captures weak local source support. Their normalized scores are later aggregated through the hierarchical fusion procedure described in Section~\ref{subsec:fusion}.

\paragraph{Entropy-weighted OT.}
A target point may have a moderate residual but an uncertain transport assignment distributed across many source-normal prototypes. We capture this using the entropy of the row-normalized transport weights:
\[
H_i
=
-\sum_{j=1}^{m}
\pi_{ij}\log(\pi_{ij}+\epsilon_0),
\]
where \(\epsilon_0\) is a small numerical constant. We use the normalized entropy
\[
\bar{H}_i = \frac{H_i}{\log m}
\]
and define
\begin{equation}
\mathcal{A}_{\mathrm{OT,ent}}^{(l)}(x_i^T)
=
\|z_i^T-\hat{z}_i\|_2
\left(
1+\eta \bar{H}_i
\right).
\label{eq:entropy_ot}
\end{equation}
where \(\eta \geq 0\) controls the influence of transport uncertainty on the anomaly score. Larger values of \(\eta\) assign more weight to diffuse, high-entropy alignments, while \(\eta=0\) reduces the score to the baseline barycentric residual.
High entropy indicates diffuse and ambiguous alignment, which increases the anomaly score. 

\paragraph{Angle-aware OT.}
Some target processes may have moderate Euclidean residual but strong directional deviation from their transported source-normal projection. We therefore define
\begin{equation}
\mathcal{A}_{\mathrm{OT,ang}}^{(l)}(x_i^T)
=
\|z_i^T-\hat{z}_i\|_2
+
\lambda
\left(
1-\cos(z_i^T,\hat{z}_i)
\right).
\label{eq:angle_ot}
\end{equation}
where \(\lambda \geq 0\) controls the contribution of angular deviation relative to the Euclidean barycentric residual. Larger values of \(\lambda\) make the score more sensitive to directional semantic drift, while \(\lambda=0\) reduces the score to the baseline residual.
This variant captures directional semantic drift between a target process and its barycentric source-normal explanation.

\paragraph{Density-aware OT.}
A target process may project close to the source-normal support but into a sparse or weakly supported region. Let
\[
d_k(\hat{z}_i)
=
\frac{1}{k}
\sum_{r \in \mathcal{N}_k(\hat{z}_i)}
\|\hat{z}_i-z_r^S\|_2
\]
be the average distance from the barycentric projection to its \(k\) nearest source-normal neighbors. Larger values indicate lower local support. We define the density-aware score as
\begin{equation}
\mathcal{A}_{\mathrm{OT,den}}^{(l)}(x_i^T)
=
\|z_i^T-\hat{z}_i\|_2
\left(
1+\beta
\frac{d_k(\hat{z}_i)}
{\operatorname{median}_{r}(d_k(z_r^S))+\epsilon_0}
\right).
\label{eq:density_ot}
\end{equation}
where \(\beta \geq 0\) controls the influence of local source-density on the anomaly score. Larger values of \(\beta\) make the score more sensitive to projections falling in sparse source-normal regions, while \(\beta=0\) reduces the score to the baseline barycentric residual. The constant \(\epsilon_0>0\) prevents division by zero.
This formulation increases the score when the barycentric projection falls in a sparse source-normal region.

\paragraph{Parameter setting.}
The entropic OT regularization parameter \(\varepsilon\) controls the smoothness of the transport plan. Unless otherwise stated, we use \(\varepsilon=0.05\), which balances transport sharpness and numerical stability. The coefficients \(\eta\), \(\lambda\), and \(\beta\) control the relative influence of entropy uncertainty, angular deviation, and local sparsity in the OT score variants, respectively. They are not learned from target labels and are fixed before evaluation. To make these coefficients interpretable, we use normalized quantities: entropy is divided by \(\log m\), the angular term is bounded through \(1-\cos(\cdot,\cdot)\), and the density term is normalized by the median source-neighborhood distance. Unless otherwise stated, we use \(\eta=0.2\), \(\lambda=0.1\), and \(\beta=0.5\). The values of \(\eta\) and \(\lambda\) follow stable preliminary grid checks, while \(\beta\) is introduced as a conservative default in the revised density-aware formulation so that local sparsity modulates, rather than dominates, the barycentric residual. None of these coefficients is selected using target labels. We further evaluate sensitivity to \(\varepsilon\), \(\eta\), \(\lambda\), and \(\beta\) in the robustness analysis.

For each backbone \(l\), the four OT variants produce four raw score vectors over the same target processes. Since these variants may have different numerical ranges, each variant score vector is first normalized using the label-free normalization procedure described in Section~\ref{subsec:fusion}. The normalized OT variants are then aggregated as
\begin{equation}
\mathcal{A}_{\mathrm{OT}}^{(l)}(x_i^T)
=
\max
\left\{
\tilde{\mathcal{A}}_{\mathrm{OT,res}}^{(l)}(x_i^T),
\tilde{\mathcal{A}}_{\mathrm{OT,ent}}^{(l)}(x_i^T),
\tilde{\mathcal{A}}_{\mathrm{OT,ang}}^{(l)}(x_i^T),
\tilde{\mathcal{A}}_{\mathrm{OT,den}}^{(l)}(x_i^T)
\right\}.
\label{eq:ot_variant_fusion}
\end{equation}

\subsection{Score Normalization and Hierarchical Fusion}
\label{subsec:fusion}

The semantic, structural, and OT scores may have different numerical scales. Before any max-based aggregation, each score vector is normalized without using labels. For a score vector \(\mathbf{a} = (a_1,\ldots,a_n)\) computed over target processes for a fixed transfer pair, path, backbone, and variant, we apply z-score normalization:
\begin{equation}
\tilde{a}_i
=
\frac{a_i-\mu_{\mathbf{a}}}
{\sigma_{\mathbf{a}}+\epsilon_0},
\label{eq:zscore_norm}
\end{equation}
where \(\mu_{\mathbf{a}}\) and \(\sigma_{\mathbf{a}}\) are computed over the unlabeled target-score distribution. This normalization uses no target labels.

Fusion is performed hierarchically. First, OT variants are fused within each backbone using Eq.~\eqref{eq:ot_variant_fusion}. Second, scores are fused across language-model backbones:
\begin{equation}
\mathcal{A}_{\mathrm{sem}}(x)
=
\max_l
\tilde{\mathcal{A}}_{\mathrm{sem}}^{(l)}(x),
\label{eq:sem_fusion}
\end{equation}
\begin{equation}
\mathcal{A}_{\mathrm{str}}(x)
=
\max_l
\tilde{\mathcal{A}}_{\mathrm{str}}^{(l)}(x),
\label{eq:str_fusion}
\end{equation}
\begin{equation}
\mathcal{A}_{\mathrm{OT}}(x)
=
\max_l
\tilde{\mathcal{A}}_{\mathrm{OT}}^{(l)}(x).
\label{eq:ot_fusion}
\end{equation}
Finally, the path-level scores are fused as
\begin{equation}
\mathcal{A}_{\mathrm{final}}(x)
=
\max
\left\{
\mathcal{A}_{\mathrm{sem}}(x),
\mathcal{A}_{\mathrm{str}}(x),
\mathcal{A}_{\mathrm{OT}}(x)
\right\}.
\label{eq:final_fusion}
\end{equation}

Max-fusion is used as the default because it is calibration-free and consistent with security triage, where a process should be prioritized if any independent evidence channel provides strong abnormality evidence. In the absence of target labels, learned weighted fusion cannot be reliably calibrated, and averaging may dilute attacks that are visible only through one modality. We nevertheless evaluate mean fusion as a representative calibration-free alternative in the experimental section.

\subsection{Algorithm}
\label{subsec:algorithm}

Algorithm~\ref{alg:source_only_cross_os} summarizes the complete source-only workflow, from source-normal memory construction to semantic, structural, and OT-based scoring, followed by label-free normalization and hierarchical fusion.

\begin{algorithm}[t!]
\caption{Source-Only Cross-OS APT Detection}
\label{alg:source_only_cross_os}

\KwIn{
Source data $\mathcal{D}_S=\{(x_i^S,y_i^S)\}$;
target data $\mathcal{D}_T=\{x_j^T\}$;
encoders $\{g_l\}_{l=1}^{L}$;
translation dictionary or prompts;
OT parameters $(\varepsilon,\eta,\lambda,\beta)$.
}
\KwOut{
Final anomaly scores $\{\mathcal{A}_{\mathrm{final}}(x_j^T)\}_{j=1}^{n_T}$.
}

Extract source-normal subset
$\mathcal{D}_S^0=\{x_i^S \mid y_i^S=0\}$\;

\ForEach{encoder $g_l$}{
    \textbf{Semantic representation}\;

    \ForEach{$x_i^S \in \mathcal{D}_S^0$}{
        Construct source behavioral sentence $s(x_i^S)$\;
        Optionally construct target-aligned sentence $\tilde{s}(x_i^S)$\;
        Add $g_l(s(x_i^S))$ and $g_l(\tilde{s}(x_i^S))$
        to source-normal memory $\mathcal{M}_S^{0,l}$\;
    }

    \ForEach{$x_j^T \in \mathcal{D}_T$}{
        Construct target behavioral sentence $s(x_j^T)$\;
        Compute target embedding $z_j^{T,l}=g_l(s(x_j^T))$\;
        Compute semantic score
        $\mathcal{A}_{\mathrm{sem}}^{(l)}(x_j^T)$ using
        Eq.~\eqref{eq:semantic_score}\;
    }

    \textbf{Structural scoring}\;
    Construct source-normal $k$-NN graph $G_S^{0,l}$ from
    $\mathcal{M}_S^{0,l}$\;
    Train VGAE on $G_S^{0,l}$\;
    Insert target nodes by connecting them to nearest source-normal neighbors\;
    Compute structural scores
    $\mathcal{A}_{\mathrm{str}}^{(l)}(x_j^T)$ using
    Eq.~\eqref{eq:structural_score}\;

    \textbf{OT scoring}\;
    Compute cost matrix between target embeddings and source-normal memory\;
    Solve entropic OT using Eq.~\eqref{eq:entropic_ot}\;
    Compute barycentric projections using
    Eq.~\eqref{eq:barycentric_projection}\;
    Compute residual, entropy-weighted, angle-aware, and density-aware OT scores\;
    Normalize and fuse OT variants using
    Eqs.~\eqref{eq:zscore_norm} and~\eqref{eq:ot_variant_fusion}\;
}

Normalize each score vector without using labels using Eq.~\eqref{eq:zscore_norm}\;
Fuse scores across encoders and evidence channels using
Eqs.~\eqref{eq:sem_fusion}--\eqref{eq:final_fusion}\;

\KwRet{$\{\mathcal{A}_{\mathrm{final}}(x_j^T)\}_{j=1}^{n_T}$}\;

\end{algorithm}

\section{Experiments}
\label{sec:experiments}

\subsection{Dataset and Setup}
\label{subsec:dataset_setup}

We evaluate the proposed framework on the DARPA Transparent Computing (TC) datasets\footnote{\url{https://www.darpa.mil/program/transparent-computing}}, a large-scale public corpus of system-level provenance traces designed for advanced threat detection \cite{BerradaCBMMTW20}\footnote{\url{https://gitlab.com/adaptdata}}. The TC corpus contains detailed execution traces collected from four operating systems: Linux, Windows, Android, and BSD \cite{DBLP:journals/fgcs/BenabderrahmaneHVCR24}. Each trace records system-level activities such as process events, file operations, process interactions, and network flows during benign execution and APT-style attack scenarios.

For each operating system, we consider two attack scenarios involving multi-stage adversarial behavior, including actions such as privilege escalation, suspicious execution, network communication, and lateral movement. The presence of four operating systems enables ordered cross-OS transfer evaluation: for each scenario, we evaluate all \(4 \times 3 = 12\) source-target transfer pairs, such as Linux\(\rightarrow\)Windows, Windows\(\rightarrow\)BSD, and Android\(\rightarrow\)Linux. This makes the dataset well suited for studying source-only cross-OS APT detection under heterogeneous system behavior.

Table~\ref{tab:class_imbalance} summarizes the class imbalance across operating systems and attack scenarios. The number of anomalous processes is extremely small relative to the total number of processes in most cases, reflecting the operational challenge of APT detection: malicious behavior is rare, sparse, and easily dominated by benign system activity.

\begin{table}[t]
\scriptsize
\centering
\caption{Class imbalance across operating systems and attack scenarios in the DARPA TC dataset. Values indicate the number of anomalous processes over the total number of processes.}
\label{tab:class_imbalance}
\begin{tabular}{lcc}
\hline
\textbf{Operating System} & \textbf{Attack Scenario 1} & \textbf{Attack Scenario 2} \\
 & \textbf{Anomalies / Total} & \textbf{Anomalies / Total} \\
\hline
\textbf{BSD}     & 13 / 75,903   & 11 / 224,624 \\
\textbf{Linux}   & 8 / 17,569    & 8 / 11,151   \\
\textbf{Windows} & 25 / 247,160  & 46 / 282,104 \\
\textbf{Android} & 9 / 102       & 13 / 12,106  \\
\hline
\end{tabular}
\end{table}

\subsection{Source-Only Evaluation Protocol}
\label{subsec:source_only_protocol}

For each ordered source-target pair, the source operating system provides labeled provenance data, while the target operating system is treated as fully unlabeled during model construction and scoring. Source labels are used only to identify source-normal processes and construct the source-normal prototype memory. Target labels are never used for representation learning, translation, graph construction, OT computation, score normalization, fusion, model selection, or hyperparameter tuning. They are used only after scoring to compute evaluation metrics. Table~\ref{tab:source_only_protocol} summarizes what information is used at each stage of the pipeline. This protocol ensures that all reported results correspond to the intended source-only cross-OS setting.

\begin{table*}[h!]
\scriptsize
\centering
\caption{Source-only cross-OS evaluation protocol. Target labels are used only for final offline evaluation.}
\label{tab:source_only_protocol}
\begin{tabular}{p{3.2cm}ccc}
\hline
\textbf{Pipeline stage} & \textbf{Uses source labels?} & \textbf{Uses target samples?} & \textbf{Uses target labels?} \\
\hline
Source-normal memory construction & Yes & No & No \\
Behavioral sentence construction & No & Yes & No \\
Token/routine translation cache & No & No & No \\
Target embedding computation & No & Yes & No \\
VGAE training & Yes, normal subset only & No & No \\
VGAE target scoring & No & Yes & No \\
OT coupling and barycentric projection & No & Yes & No \\
Score normalization and fusion & No & Yes & No \\
AUC and nDCG computation & No & Yes & Yes, evaluation only \\
\hline
\end{tabular}
\end{table*}

\subsection{Reproducibility}
\label{subsec:reproducibility}

All experiments were conducted locally on a MacBook Pro equipped with an Apple M1 Max chip and 64 GB of unified memory, running macOS 15 (Sequoia). 

\subsection{Implementation Details}
\label{subsec:implementation_details}

We evaluate four pretrained language-model encoders: BERT \cite{devlin2019bert}, RoBERTa \cite{liu2019roberta}, DistilBERT \cite{sanh2019distilbert}, and MiniLM \cite{wang2020minilm}. MiniLM embeddings are \(384\)-dimensional, while BERT, RoBERTa, and DistilBERT embeddings are \(768\)-dimensional. For each process, structured behavioral sentences are generated from provenance logs using deterministic templates. OS-specific tokens, such as routines, executable names, and event identifiers, are optionally canonicalized through restricted prompt-based mapping. To control cost and runtime, translations are cached at the token/routine level rather than at the process level; therefore, the number of API calls grows with the number of unique technical tokens rather than with the number of processes.

For semantic scoring, each target process is compared against the source-normal prototype memory using minimum cosine distance. For structural scoring, we construct a \(k\)-nearest-neighbor graph over source-normal prototypes for each embedding backbone and train a VGAE on the source-normal graph. Unless otherwise stated, the neighborhood size is defined as\\ \(k=\max(k_{\min}, \lceil \rho N_S^{0,l}\rceil)\), where \(N_S^{0,l}\) is the size of the source-normal prototype memory for encoder \(l\). For OT scoring, we compute entropic OT between target embeddings and the source-normal prototype memory using the POT library. We use \(\varepsilon=0.05\) as the default entropic regularization strength.

For the OT score variants, we use \(\eta=0.2\) for entropy-weighted OT, \(\lambda=0.1\) for angle-aware OT, and \(\beta=0.5\) for density-aware OT. These coefficients are fixed before evaluation and are not selected using target labels. Score vectors are normalized using z-score normalization over the unlabeled target-score distribution before max-based aggregation. This normalization is applied separately for each transfer pair, scoring path, embedding backbone, and OT variant, following Eq.~\eqref{eq:zscore_norm}. We further evaluate sensitivity to these hyperparameters in the robustness analysis.

\subsection{Comparison With Existing Methods}
\label{subsec:baselines}

A large body of work studies APT detection within a single operating system using provenance graphs, endpoint telemetry, or environment-specific behavioral representations \cite{Flash24,DBLP:journals/fgcs/BenabderrahmaneHVCR24}. Such systems are often highly effective in their intended deployment setting, but they are not designed for the protocol studied in this paper: source-only cross-OS transfer with no target-domain supervision. In particular, many within-OS detectors rely on target-specific feature engineering, environment-dependent training, or labeled attack traces in the deployment domain. Directly comparing against such supervised target-domain methods would violate the zero-target-label premise of our setting.

We therefore compare against source-only transfer baselines that can be trained using source-domain data and applied directly to the unlabeled target domain. These baselines are selected because they satisfy the same information constraints as our method: no target labels are used for training, calibration, fusion, or model selection.

\begin{itemize}
    \item \textbf{AE Transfer.} An autoencoder is trained on source-normal embeddings and applied to target embeddings. The reconstruction error is used as the anomaly score \cite{boone2025joint}.

    \item \textbf{Isolation Forest Transfer.} Isolation Forest is trained on source-normal embeddings and then applied to target embeddings. Samples that are easier to isolate receive higher anomaly scores \cite{xue2025novel}.

    \item \textbf{\(k\)-NN Transfer.} Each target sample is scored by its distance to the \(k\) nearest source-normal embeddings. Larger distances indicate weaker support from the source-normal reference \cite{sorkhei2025k}.

    \item \textbf{LOF Transfer.} Local Outlier Factor is fit using source-normal embeddings and applied to target embeddings to detect deviations from the local density structure of the source domain \cite{vincent2020transfer}.
\end{itemize}

These baselines provide protocol-aligned references for assessing whether the proposed semantic, structural, and OT-based components improve cross-OS generalization beyond standard source-only anomaly scoring. For all baselines, the training reference is restricted to source-normal data, and target labels are used only for final evaluation.

\subsection{Evaluation Metrics}
\label{subsec:metrics}

We evaluate anomaly detection performance using both ROC-AUC and nDCG. These metrics capture complementary aspects of the source-only cross-OS detection problem.

\paragraph{ROC-AUC}
ROC-AUC measures the threshold-independent ability of a detector to rank anomalous processes above benign processes. This is useful because the operating threshold is typically unknown in advance and may vary across SOC deployments. ROC-AUC therefore captures global separability between benign and attack-related processes across the entire ranked list.

\paragraph{nDCG}
Because APT detection is an extreme class-imbalance problem, global separability alone is not sufficient. In practice, analysts usually inspect only the highest-ranked alerts. We therefore also report Normalized Discounted Cumulative Gain (nDCG), a ranking-oriented metric that rewards placing true anomalies near the top of the alert list \cite{jarvelin2002cumulated,yang2025breaking}. Given a ranked list of target processes, nDCG is defined as
\[
\mathrm{nDCG@}K =
\frac{\mathrm{DCG@}K}{\mathrm{IDCG@}K},
\]
where
\[
\mathrm{DCG@}K =
\sum_{i=1}^{K}
\frac{2^{rel_i}-1}{\log_2(i+1)}.
\]
Here, \(rel_i\in\{0,1\}\) indicates whether the process ranked at position \(i\) is anomalous, and \(\mathrm{IDCG@}K\) denotes the ideal DCG obtained by a perfect ranking. A value of \(1\) indicates that all relevant anomalous processes are ranked as early as possible. Unless otherwise stated, we report nDCG over the full ranked target list.
\subsection{Experimental Organization}
\label{subsec:evaluation_protocol}

We evaluate the proposed framework under a strict source-only cross-OS protocol. For each of the two attack scenarios, we consider all ordered source-target operating-system pairs among Linux, Windows, BSD, and Android. This results in \(12\) transfer pairs per scenario and \(24\) transfer settings in total. In each setting, the source domain is used to construct the source-normal reference and train source-only components, while the target domain is treated as unlabeled throughout representation construction, translation, graph construction, OT computation, scoring, normalization, and fusion. Target labels are used only after scoring, for offline computation of ROC-AUC and nDCG.

The evaluation is organized around six complementary analyses. First, we compare the proposed label-free fused framework against protocol-aligned source-only baselines. The proposed score is computed by fusing normalized per-process semantic, structural, and OT-based anomaly scores before evaluation. The dominant evidence channel is reported only as a diagnostic interpretation of which component most strongly explains the result in each transfer direction; it is not used for supervised model selection or target-domain tuning.

Second, we compare against four source-only transfer baselines: AE Transfer, Isolation Forest Transfer, \(k\)-NN Transfer, and LOF Transfer. Each baseline is trained using only source-normal embeddings and then applied directly to target-domain embeddings without using target labels. This ensures that all methods operate under the same source-only information constraint. We report both per-transfer results and aggregate summaries over the \(12\) transfer pairs of each scenario.

Third, we perform score-level component ablation to quantify the contribution of each evidence channel under the actual deployable fusion rule. We evaluate semantic-only, structural-only, OT-only, pairwise combinations, and the complete semantic+structural+OT configuration.

Fourth, we evaluate the fusion strategy itself. The default max-fusion rule is compared with mean fusion under the same label-free normalization procedure. This analysis tests whether preserving the strongest anomaly evidence from any single view is preferable to averaging evidence across views. Both fusion strategies operate without target labels.

Fifth, we analyze the four OT scoring variants: residual OT, entropy-weighted OT, angle-aware OT, and density-aware OT. This analysis is performed across all transfer pairs and all four embedding backbones: BERT, RoBERTa, DistilBERT, and MiniLM. The goal is not to assume that one OT variant dominates universally, but to evaluate whether the variants capture complementary forms of cross-OS geometric mismatch.

Finally, we study cross-OS transfer asymmetry and robustness. We examine the complementarity of semantic, structural, and geometric evidence by reporting which component is dominant across transfer directions. We also visualize source-target transfer behavior using heatmaps, where rows correspond to source operating systems and columns correspond to target operating systems, with diagonal self-transfer entries omitted. This representation highlights cross-OS asymmetry and makes target-domain effects easier to interpret. In addition, we assess sensitivity to key OT hyperparameters to verify that the observed trends are stable under reasonable perturbations and are not driven by an isolated parameter choice.

Across four language-model backbones, three evidence channels, four OT variants, four source-only baselines, twelve transfer pairs, and two attack scenarios, the evaluation covers a broad set of source-only cross-OS configurations. For stochastic components such as VGAE training, experiments are repeated over multiple random seeds when applicable. For deterministic scoring components, we report averages across transfer pairs and scenarios, and we use the full per-transfer results to assess stability across heterogeneous source-target directions.

To ensure fair comparison, all methods receive the same source and target embeddings for a given backbone and transfer pair. No method is allowed to use target labels for training, hyperparameter selection, score calibration, or fusion. When a method requires a normal reference, it is trained only on source-normal embeddings. Score normalization is performed using the same label-free procedure before aggregation whenever required.
\subsection{Controlled Cross-OS Keyword Translation Implementation}
\label{subsec:controlled_translation}

A major challenge in source-only cross-OS transfer is that the same behavioral intent may be expressed through different operating-system vocabularies. For example, a Linux process may execute \texttt{/bin/bash}, load a shared object such as \texttt{lib.so}, or copy a shell payload, whereas an analogous Windows behavior may involve \texttt{cmd.exe}, executable or library-style artifacts, and Windows-style file paths. Directly embedding these raw OS-specific tokens can increase the apparent semantic distance between otherwise related behaviors. To reduce this vocabulary mismatch, we apply a controlled keyword-level translation step before embedding. Importantly, the translation module is not allowed to rewrite the full behavioral trace freely. Instead, it receives a list of technical tokens extracted from the source sentence and is instructed to map only OS-specific identifiers to target-domain equivalents when a reliable conceptual match exists. Observed actions, event types, process identifiers, IP addresses, ports, and the order of events are preserved. If no reliable equivalent exists, the token is retained or mapped to a generic behavioral description. This design reduces hallucination risk because the language model is used as a constrained vocabulary canonicalizer rather than as a generative event synthesizer. For example, consider the following Linux source sentence:
\begin{quote}
\small
\texttt{Process 42 executed /bin/bash, loaded /usr/lib/lib.so, copied /tmp/backdoor/malware\_payload.sh to /home/user/, opened /etc/passwd, and sent data through ssh to 192.168.1.10:22.}
\end{quote}

For a Linux\(\rightarrow\)Windows transfer setting, the controlled translation step may produce:
\begin{quote}
\small
\texttt{Process 42 executed cmd.exe,}\\
\texttt{loaded C:/Windows/System32/lib.exe,}\\
\texttt{copied C:/tmp/backdoor/malware\_payload.exe}\\
\texttt{to C:/Users/user/,}\\
\texttt{accessed an account-information file,}\\
\texttt{and sent data through an SSH client to 192.168.1.10:22.}
\end{quote}

The goal is not to assert that the exact Windows artifacts were observed. Rather, the translated sentence expresses the same source-domain behavior using target-domain terminology when an equivalent concept exists. Thus, \texttt{/bin/bash} may be mapped to \texttt{cmd.exe}, a Linux shared object such as \texttt{lib.so} may be mapped to a Windows-style executable or library artifact such as \texttt{lib.exe}, and a Linux home-directory path may be rewritten using a Windows user-directory convention. Network endpoints and event semantics are preserved.

The prompt used for this step is restricted to token-level mapping. A representative prompt is shown below:
\begin{quote}
\small
\texttt{You are given a list of technical tokens extracted from a Linux provenance sentence. Translate only OS-specific keywords into Windows-equivalent terminology when a clear conceptual equivalent exists. Do not add new events, processes, files, IP addresses, ports, or behaviors. Preserve the original action sequence. If no reliable equivalent exists, keep the token unchanged or replace it with a generic behavioral description. Return only a JSON dictionary mapping source tokens to translated tokens.}
\end{quote}

For the example above, the model may return a dictionary such as:
\[
\begin{aligned}
\texttt{/bin/bash} &\rightarrow \texttt{cmd.exe},\\
\texttt{/usr/lib/lib.so} &\rightarrow \texttt{C:/Windows/System32/lib.exe},\\
\texttt{malware\_payload.sh} &\rightarrow \texttt{malware\_payload.exe},\\
\texttt{/home/user/} &\rightarrow \texttt{C:/Users/user/},\\
\texttt{/etc/passwd} &\rightarrow \texttt{account-information file}.
\end{aligned}
\]

To reduce computation cost and improve consistency, we cache translations at the token and routine level. Many processes share the same technical identifiers, such as executable names, library names, event labels, or routine names. Therefore, once a token such as \texttt{/bin/bash} or \texttt{lib.so} has been mapped for a given source-target OS pair, the same cached mapping is reused for all subsequent processes. This avoids repeated LLM calls, ensures consistent canonicalization across the dataset, and makes the number of translation queries depend on the size of the unique technical vocabulary rather than on the number of processes.

This controlled translation strategy preserves the source-only evaluation protocol. Target labels are never used during translation, embedding, scoring, normalization, or fusion. The translation step only aligns technical vocabulary between operating systems so that semantically related behaviors can be compared in a shared embedding space.

\section{Results and Analysis}
\label{sec:results_analysis}

We organize the results around six questions:
(i) whether the proposed source-only framework improves over protocol-aligned baselines;
(ii) whether score-level fusion improves over individual evidence channels;
(iii) whether semantic, structural, and OT-based evidence provide complementary signals;
(iv) how the individual OT variants behave across transfer directions and embedding backbones;
(v) how performance depends on the source-target operating-system pair; and
(vi) whether the main hyperparameter choices are stable under reasonable perturbations.
Unless otherwise stated, entries of the form \(a/b\) denote ROC-AUC/nDCG. The gain column reports absolute ROC-AUC improvement over the strongest source-only baseline among AE, \(k\)-NN, IF, and LOF. All fusion results are computed without target labels; target labels are used only for offline evaluation.

\subsection{Comparison with Source-Only Transfer Baselines}
\label{subsec:baseline_comparison}

This analysis evaluates whether the proposed source-only framework improves over protocol-aligned transfer baselines under the same no-target-label constraint. The proposed score is obtained through label-free score-level fusion of semantic, structural, and OT-based evidence. The dominant evidence column is included only for interpretation; it indicates which evidence channel most strongly explains the result in the diagnostic component analysis and is not used for supervised model selection.

\begin{table*}[h!]
\centering
\scriptsize
\caption{
Scenario~1 source-only baseline comparison using ROC-AUC/nDCG.
The Proposed column reports the score produced by the proposed label-free framework.
Dominant Evidence indicates the component that most strongly explains the result in diagnostic analysis.
Gain reports absolute ROC-AUC improvement over the strongest source-only baseline.
}
\label{tab:scenario1_component_synthesis}
\begin{adjustbox}{max width=\textwidth}
\small
\begin{tabular}{lccccccc}
\toprule
\textbf{Transfer Pair}
& \textbf{Proposed}
& \textbf{Dominant Evidence}
& \textbf{AE}
& \textbf{\(k\)-NN}
& \textbf{IF}
& \textbf{LOF}
& \textbf{Gain} \\
\midrule
Linux \(\rightarrow\) Windows   & \textbf{0.79/0.60} & OT & 0.52/0.29 & 0.55/0.31 & 0.37/0.18 & 0.60/0.41 & +0.19 \\
Linux \(\rightarrow\) BSD       & \textbf{0.98/0.61} & OT & 0.69/0.51 & 0.78/0.54 & 0.57/0.38 & 0.70/0.49 & +0.20 \\
Linux \(\rightarrow\) Android   & \textbf{0.93/0.68} & OT & 0.75/0.52 & 0.81/0.55 & 0.61/0.49 & 0.78/0.55 & +0.12 \\
BSD \(\rightarrow\) Windows     & \textbf{0.81/0.63} & Str & 0.19/0.09 & 0.45/0.29 & 0.16/0.12 & 0.42/0.22 & +0.36 \\
BSD \(\rightarrow\) Linux       & \textbf{0.98/0.57} & Sem/OT & 0.71/0.49 & 0.67/0.41 & 0.59/0.39 & 0.81/0.42 & +0.17 \\
BSD \(\rightarrow\) Android     & \textbf{0.81/0.70} & OT & 0.45/0.38 & 0.36/0.22 & 0.33/0.19 & 0.56/0.40 & +0.25 \\
Windows \(\rightarrow\) BSD     & \textbf{0.79/0.59} & OT & 0.53/0.40 & 0.57/0.40 & 0.57/0.39 & 0.67/0.46 & +0.12 \\
Windows \(\rightarrow\) Linux   & \textbf{0.81/0.56} & OT & 0.66/0.46 & 0.63/0.49 & 0.58/0.41 & 0.60/0.48 & +0.15 \\
Windows \(\rightarrow\) Android & \textbf{0.80/0.71} & OT & 0.63/0.50 & 0.60/0.44 & 0.40/0.29 & 0.56/0.40 & +0.17 \\
Android \(\rightarrow\) Windows & \textbf{0.98/0.80} & OT & 0.60/0.51 & 0.71/0.53 & 0.57/0.44 & 0.66/0.59 & +0.27 \\
Android \(\rightarrow\) Linux   & \textbf{0.78/0.55} & OT & 0.57/0.41 & 0.70/0.49 & 0.52/0.38 & 0.47/0.30 & +0.08 \\
Android \(\rightarrow\) BSD     & \textbf{0.96/0.61} & OT & 0.80/0.56 & 0.58/0.47 & 0.55/0.36 & 0.46/0.28 & +0.16 \\
\bottomrule
\end{tabular}
\end{adjustbox}
\end{table*}

\begin{table*}[h!]
\centering
\scriptsize
\caption{
Scenario~2 source-only baseline comparison using ROC-AUC/nDCG.
The Proposed column reports the score produced by the proposed label-free framework.
Dominant Evidence indicates the component that most strongly explains the result in diagnostic analysis.
Gain reports absolute ROC-AUC improvement over the strongest source-only baseline.
}
\label{tab:scenario2_component_synthesis}
\begin{adjustbox}{max width=\textwidth}
\small
\begin{tabular}{lccccccc}
\toprule
\textbf{Transfer Pair}
& \textbf{Proposed}
& \textbf{Dominant Evidence}
& \textbf{AE}
& \textbf{\(k\)-NN}
& \textbf{IF}
& \textbf{LOF}
& \textbf{Gain} \\
\midrule
Linux \(\rightarrow\) Windows   & \textbf{0.78/0.38} & OT & 0.36/0.10 & 0.50/0.22 & 0.45/0.18 & 0.66/0.33 & +0.12 \\
Linux \(\rightarrow\) BSD       & \textbf{0.98/0.63} & OT & 0.67/0.30 & 0.51/0.25 & 0.80/0.45 & 0.16/0.09 & +0.18 \\
Linux \(\rightarrow\) Android   & \textbf{0.79/0.70} & OT & 0.34/0.10 & 0.36/0.12 & 0.60/0.42 & 0.61/0.25 & +0.18 \\
BSD \(\rightarrow\) Windows     & \textbf{0.76/0.38} & OT & 0.22/0.10 & 0.51/0.30 & 0.60/0.31 & 0.60/0.30 & +0.16 \\
BSD \(\rightarrow\) Linux       & \textbf{0.85/0.51} & OT & 0.63/0.41 & 0.52/0.28 & 0.52/0.19 & 0.72/0.42 & +0.13 \\
BSD \(\rightarrow\) Android     & \textbf{0.93/0.66} & OT & 0.32/0.10 & 0.56/0.17 & 0.48/0.26 & 0.66/0.39 & +0.27 \\
Windows \(\rightarrow\) BSD     & \textbf{0.97/0.59} & Sem/OT & 0.70/0.49 & 0.58/0.34 & 0.70/0.42 & 0.60/0.27 & +0.27 \\
Windows \(\rightarrow\) Linux   & \textbf{0.92/0.65} & Str & 0.30/0.12 & 0.56/0.20 & 0.22/0.10 & 0.45/0.13 & +0.36 \\
Windows \(\rightarrow\) Android & \textbf{0.79/0.70} & OT & 0.65/0.33 & 0.54/0.21 & 0.32/0.10 & 0.63/0.20 & +0.14 \\
Android \(\rightarrow\) Windows & \textbf{0.75/0.51} & Str & 0.60/0.38 & 0.61/0.30 & 0.56/0.31 & 0.54/0.29 & +0.14 \\
Android \(\rightarrow\) Linux   & \textbf{0.91/0.60} & OT & 0.32/0.10 & 0.56/0.30 & 0.58/0.35 & 0.60/0.33 & +0.31 \\
Android \(\rightarrow\) BSD     & \textbf{0.92/0.70} & Sem/Str/OT & 0.55/0.34 & 0.22/0.18 & 0.64/0.35 & 0.56/0.28 & +0.28 \\
\bottomrule
\end{tabular}
\end{adjustbox}
\end{table*}

Tables~\ref{tab:scenario1_component_synthesis} and~\ref{tab:scenario2_component_synthesis} show that the proposed framework improves over the strongest source-only baseline in every source-target transfer direction for both attack scenarios. In Scenario~1, the ROC-AUC gain ranges from \(+0.08\) to \(+0.36\), while in Scenario~2 it ranges from \(+0.12\) to \(+0.36\). The nDCG values show the same operational trend: the proposed framework improves not only global separability but also top-ranked alert quality. This is important for APT triage, where analysts typically inspect only a small set of high-priority processes.

\subsection{Aggregate Performance Across Scenarios}
\label{subsec:aggregate_summary}

This analysis summarizes the main quantitative trends across all 24 source-target transfer settings, covering 12 transfer pairs in each of the two attack scenarios.

\begin{table}[h!]
\centering
\scriptsize
\caption{
Aggregate comparison over the 12 ordered cross-OS transfer pairs in each scenario.
Values report mean ROC-AUC/nDCG.
}
\label{tab:baseline_aggregate_summary}
\begin{tabular}{lccc}
\toprule
\textbf{Method} & \textbf{Scenario 1} & \textbf{Scenario 2} & \textbf{Overall} \\
\midrule
AE Transfer        & 0.59/0.42 & 0.47/0.23 & 0.53/0.33 \\
\(k\)-NN Transfer  & 0.61/0.42 & 0.50/0.23 & 0.56/0.33 \\
IF Transfer        & 0.48/0.33 & 0.53/0.28 & 0.51/0.31 \\
LOF Transfer       & 0.61/0.41 & 0.56/0.27 & 0.59/0.35 \\
Proposed           & \textbf{0.86/0.63} & \textbf{0.86/0.58} & \textbf{0.86/0.60} \\
\bottomrule
\end{tabular}
\end{table}

Table~\ref{tab:baseline_aggregate_summary} confirms the consistency of the gains. The strongest baseline in terms of overall mean ROC-AUC is LOF with \(0.59\), whereas the proposed framework reaches \(0.86\). The same pattern holds for nDCG: the strongest baseline reaches \(0.35\), while the proposed framework reaches \(0.60\). These results support the use of semantic abstraction, structural modeling, and OT-based geometric scoring for source-only cross-OS APT detection.

\subsection{Score-Level Component Ablation}
\label{subsec:score_level_ablation}

This ablation evaluates the contribution of each evidence channel under the actual score-level fusion rule. Scores are normalized and combined before computing ROC-AUC, so this experiment reflects the deployable label-free fusion procedure.

\begin{table}[h!]
\centering
\scriptsize
\caption{
Score-level component ablation using label-free fusion. Values report mean ROC-AUC over the 12 ordered cross-OS transfer pairs for each scenario.
}
\label{tab:score_level_ablation}
\begin{tabular}{lcc}
\toprule
\textbf{Configuration} & \textbf{Scenario 1} & \textbf{Scenario 2} \\
\midrule
Semantic only              & 0.742 & 0.554 \\
Structural only            & 0.803 & 0.815 \\
OT only                    & 0.855 & 0.811 \\
Semantic + Structural      & 0.828 & 0.832 \\
Semantic + OT              & 0.860 & 0.824 \\
Structural + OT            & 0.862 & 0.841 \\
Semantic + Structural + OT & \textbf{0.864} & \textbf{0.858} \\
\bottomrule
\end{tabular}
\end{table}

Table~\ref{tab:score_level_ablation} shows that the complete three-path fusion achieves the highest mean ROC-AUC in both scenarios. OT is the strongest individual component in Scenario~1, while structural and OT evidence are both highly competitive in Scenario~2. Pairwise combinations improve over most single components, and the full fusion provides the best overall result. This confirms that the framework benefits from combining complementary evidence channels under the actual label-free scoring rule.

\subsection{Fusion Strategy Ablation}
\label{subsec:fusion_ablation}

To justify the default max-fusion rule, we compare it with mean fusion under the same label-free normalization procedure. Both strategies combine normalized per-process anomaly scores and do not use target labels.

\begin{table}[h!]
\centering
\scriptsize
\caption{
Label-free fusion strategy ablation. Values report mean ROC-AUC over the 12 ordered cross-OS transfer pairs for each attack scenario.
}
\label{tab:fusion_strategy_ablation}
\begin{tabular}{lcc}
\toprule
\textbf{Fusion Strategy} & \textbf{Scenario 1} & \textbf{Scenario 2} \\
\midrule
Mean fusion & 0.75 & 0.71 \\
Max fusion  & \textbf{0.86} & \textbf{0.86} \\
\bottomrule
\end{tabular}
\end{table}

Table~\ref{tab:fusion_strategy_ablation} shows that max fusion outperforms mean fusion in both attack scenarios. This supports our design choice: in source-only cross-OS APT detection, an anomalous process may be strongly visible in only one evidence channel. Mean fusion can dilute such single-view evidence, whereas max fusion preserves the strongest semantic, structural, or geometric anomaly signal.

\subsection{Complementarity of Semantic, Structural, and OT Evidence}
\label{subsec:component_complementarity}

This analysis examines whether the semantic, structural, and OT-based evidence channels are redundant or complementary across transfer directions.

\begin{table}[h!]
\centering
\scriptsize
\caption{
Dominant evidence channel across the two attack scenarios.
Counts are computed from Tables~\ref{tab:scenario1_component_synthesis} and~\ref{tab:scenario2_component_synthesis}.
}
\label{tab:winner_counts}
\begin{tabular}{lccc}
\toprule
\textbf{Dominant evidence channel} & \textbf{Scenario 1} & \textbf{Scenario 2} & \textbf{Total} \\
\midrule
Semantic only & 0/12 & 0/12 & 0/24 \\
Structural GVAE & 1/12 & 2/12 & 3/24 \\
OT geometric evidence & 10/12 & 8/12 & 18/24 \\
Tie involving multiple channels & 1/12 & 2/12 & 3/24 \\
\bottomrule
\end{tabular}
\end{table}

Table~\ref{tab:winner_counts} shows that OT is the dominant evidence channel in most transfer directions, supporting its role as the central geometry-aware transfer mechanism. At the same time, structural evidence is essential in several difficult transfers, such as those where graph-neighborhood behavior better captures cross-OS consistency than direct semantic or geometric alignment. The ties further show that different evidence channels can converge to similarly strong rankings. This supports the design choice of retaining all three paths rather than reducing the framework to a single detector.

\subsection{OT Variant Analysis}
\label{subsec:ot_variant_analysis}

This analysis evaluates how the four OT scoring variants behave across embedding backbones and transfer directions. The goal is not to claim that one OT variant dominates universally, but to verify that the variants provide complementary geometric signals.

\begin{table*}[h!]
\centering
\scriptsize
\caption{
OT variant analysis across both attack scenarios. BERT, RoBERTa, DistilBERT, and MiniLM columns report mean ROC-AUC over the 12 transfer pairs for the corresponding backbone. Avg.\(\pm\)Std, median, worst-case, and AUC\(>0.5\) are computed over 48 configurations: 12 transfer pairs \(\times\) 4 embedding backbones. Wins/12 counts how often the variant obtains the best average AUC across the four backbones for a transfer pair; ties are counted for all tied variants.
}
\label{tab:ot_variant_analysis}
\resizebox{\textwidth}{!}{
\begin{tabular}{llccccccccc}
\toprule
\textbf{Scenario} & \textbf{OT Variant} 
& \textbf{BERT} 
& \textbf{RoBERTa} 
& \textbf{DistilBERT} 
& \textbf{MiniLM} 
& \textbf{Avg.\(\pm\)Std} 
& \textbf{Median} 
& \textbf{Worst} 
& \textbf{Wins/12} 
& \textbf{AUC\(>0.5\)} \\
\midrule
S1 & Residual OT         & 0.639 & 0.659 & 0.622 & 0.698 & \textbf{0.655\(\pm\)0.168} & 0.640 & 0.320 & 5 & 40/48 \\
S1 & Entropy-weighted OT & 0.636 & 0.652 & 0.610 & 0.707 & 0.651\(\pm\)0.165 & 0.670 & 0.320 & 2 & 40/48 \\
S1 & Angle-aware OT      & 0.616 & 0.618 & 0.605 & 0.694 & 0.633\(\pm\)0.149 & 0.675 & 0.300 & 3 & 40/48 \\
S1 & Density-aware OT    & 0.636 & 0.626 & 0.628 & 0.717 & 0.652\(\pm\)0.134 & 0.675 & 0.300 & 2 & 42/48 \\
\midrule
S2 & Residual OT         & 0.467 & 0.628 & 0.578 & 0.573 & 0.562\(\pm\)0.217 & 0.535 & 0.160 & 5 & 29/48 \\
S2 & Entropy-weighted OT & 0.511 & 0.588 & 0.604 & 0.569 & 0.568\(\pm\)0.200 & 0.610 & 0.150 & 4 & 32/48 \\
S2 & Angle-aware OT      & 0.507 & 0.582 & 0.625 & 0.622 & \textbf{0.584\(\pm\)0.250} & \textbf{0.630} & 0.120 & 2 & 30/48 \\
S2 & Density-aware OT    & 0.487 & 0.576 & 0.513 & 0.647 & 0.556\(\pm\)0.251 & 0.615 & 0.120 & 2 & 26/48 \\
\bottomrule
\end{tabular}
}
\end{table*}

Table~\ref{tab:ot_variant_analysis} shows that OT behavior is scenario-dependent. In Scenario~1, all variants are relatively stable, with most configurations exceeding random performance. In Scenario~2, the task is harder and variability increases, but the OT variants still provide useful geometric evidence. Angle-aware OT obtains the strongest mean and median AUC in Scenario~2, while residual OT wins the largest number of transfer pairs after averaging over backbones. This supports the use of normalized OT-variant fusion: each variant captures a different form of geometric mismatch, and no single variant is optimal for all source-target directions.

\subsection{Source--Target Transfer Behavior}
\label{subsec:transfer_heatmaps}

To visualize the asymmetry of cross-OS transfer, we represent component-level AUCs as source-target heatmaps. Rows correspond to source operating systems and columns correspond to target operating systems; diagonal self-transfer entries are omitted.

\begin{figure*}[t]
    \centering
    \begin{subfigure}{0.36\linewidth}
        \centering
        \includegraphics[width=\linewidth]{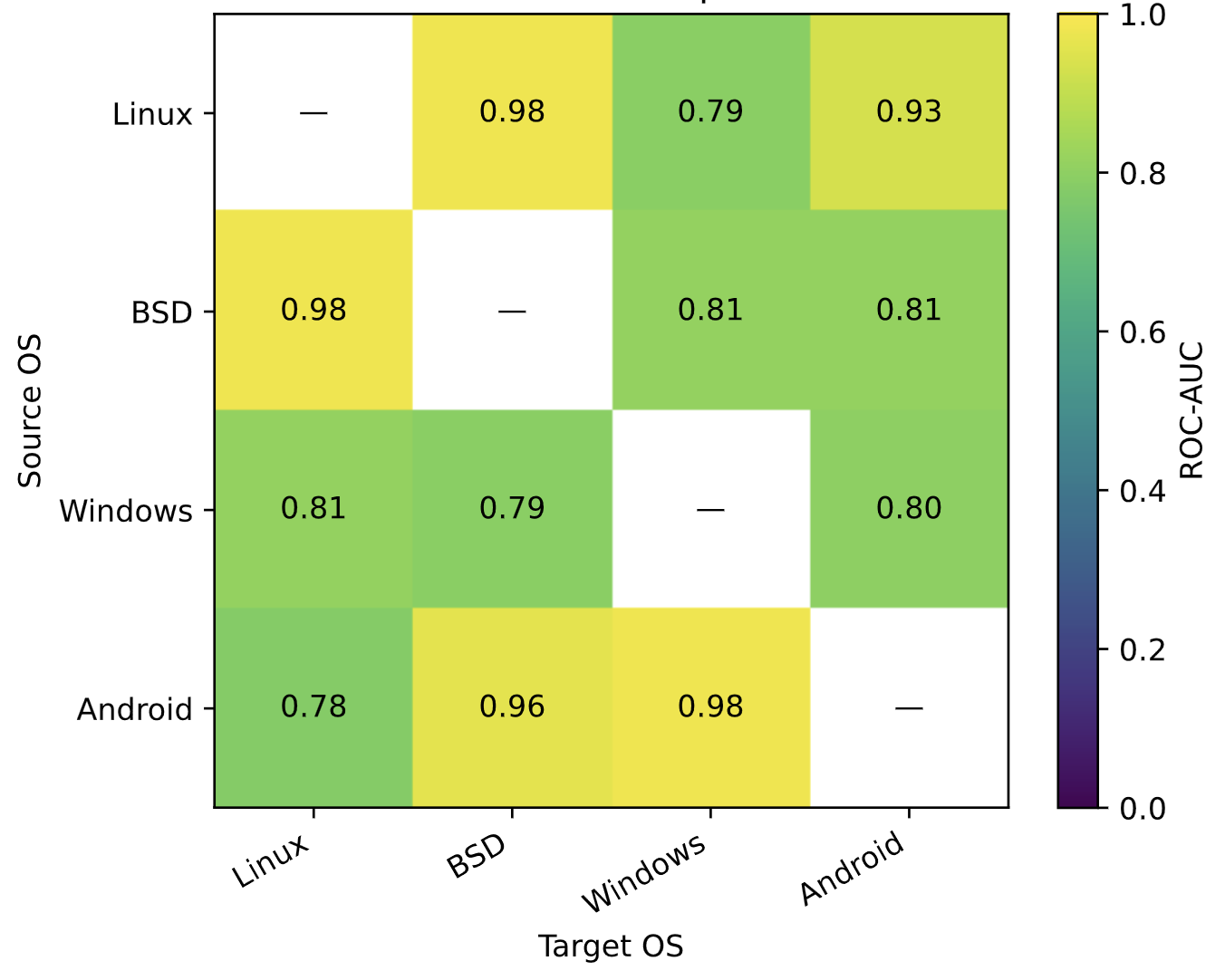}
        \caption{Scenario~1: fused framework}
    \end{subfigure}
    \hfill
    \begin{subfigure}{0.36\linewidth}
        \centering
        \includegraphics[width=\linewidth]{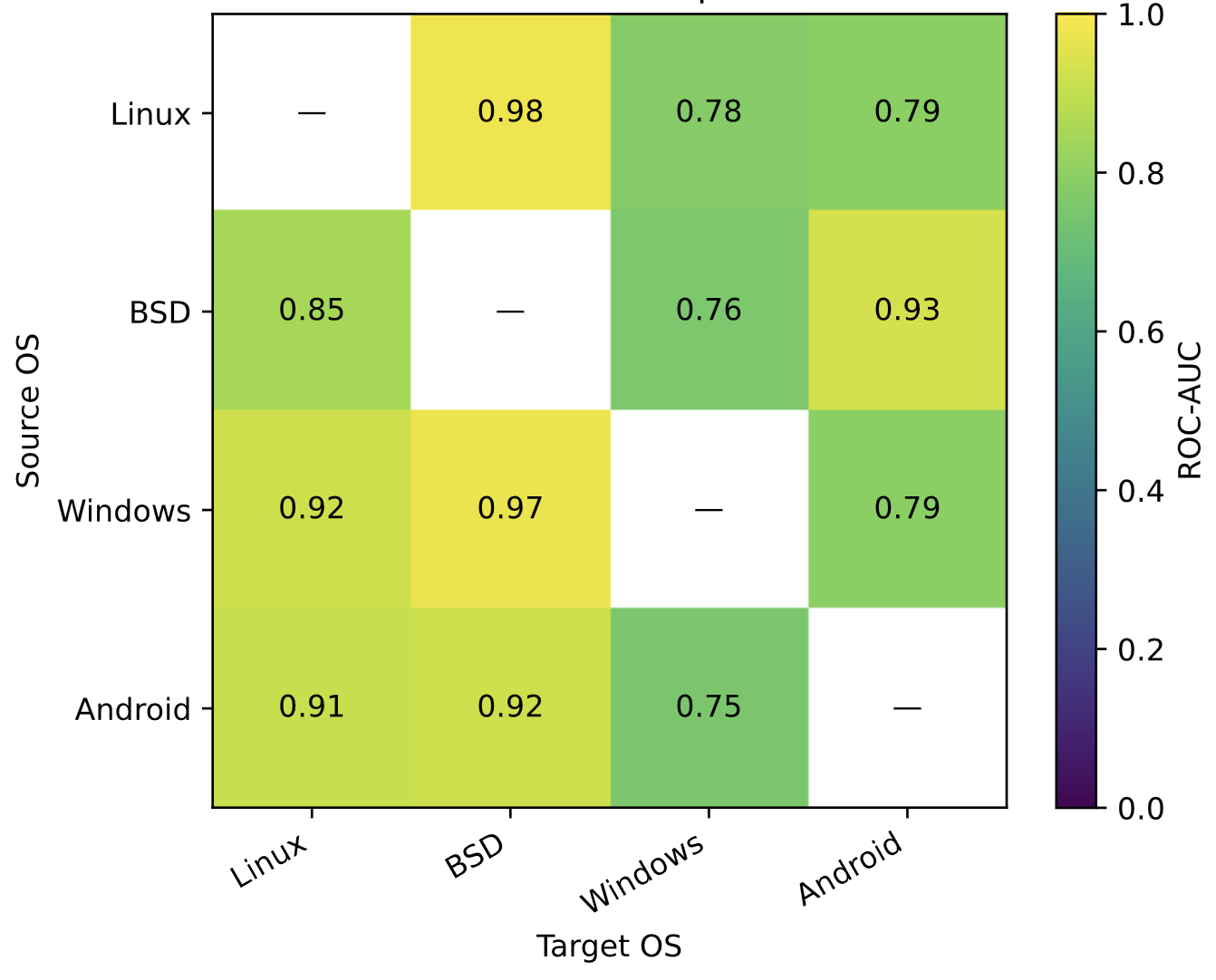}
        \caption{Scenario~2: fused framework}
    \end{subfigure}
    \caption{
    Source-target transfer heatmaps for the proposed fused framework. Rows denote the source operating system and columns denote the target operating system; diagonal self-transfer entries are omitted. The heatmaps show that the proposed source-only framework remains effective across heterogeneous cross-OS transfer directions in both attack scenarios, while also revealing asymmetric transfer difficulty across source-target pairs.
    }
    \label{fig:fused_transfer_heatmaps}
\end{figure*}

Figure~\ref{fig:fused_transfer_heatmaps} summarizes the cross-OS transfer behavior of the proposed fused framework. The results show that performance is not symmetric across source-target directions, confirming that cross-OS APT detection depends strongly on the target platform and attack scenario. Nevertheless, the fused framework maintains strong performance across most transfer pairs in both scenarios, supporting the use of complementary semantic, structural, and OT-based evidence.


\subsection{Hyperparameter Sensitivity}
\label{subsec:hyperparameter_sensitivity}

The goal of this experiment is not to tune parameters on target labels, but to verify that the observed trends are stable under reasonable perturbations around the default settings used in the main experiments.

\begin{table}[h!]
\centering
\scriptsize
\caption{
Sensitivity to OT hyperparameters. Values report mean ROC-AUC over the 12 ordered cross-OS transfer pairs. Bold values correspond to the default setting used in the main experiments.
}
\label{tab:ot_hyperparameter_sensitivity}
\begin{tabular}{lccc}
\toprule
\textbf{Parameter} & \textbf{Value} & \textbf{Scenario 1} & \textbf{Scenario 2} \\
\midrule
\(\varepsilon\) & 0.01 & 0.81 & 0.82 \\
\(\varepsilon\) & \textbf{0.05} & \textbf{0.86} & \textbf{0.86} \\
\(\varepsilon\) & 0.10 & 0.84 & 0.83 \\
\midrule
\(\eta\) & 0.10 & 0.79 & 0.78 \\
\(\eta\) & \textbf{0.20} & \textbf{0.86} & \textbf{0.86} \\
\(\eta\) & 0.50 & 0.85 & 0.83 \\
\midrule
\(\lambda\) & 0.05 & 0.77 & 0.81 \\
\(\lambda\) & \textbf{0.10} & \textbf{0.86} & \textbf{0.86} \\
\(\lambda\) & 0.20 & 0.86 & 0.85 \\
\midrule
\(\beta\) & 0.25 & 0.85 & 0.84 \\
\(\beta\) & \textbf{0.50} & \textbf{0.86} & \textbf{0.86} \\
\bottomrule
\end{tabular}
\end{table}

Table~\ref{tab:ot_hyperparameter_sensitivity} shows that the default settings provide the strongest or near-strongest performance across both scenarios. Moderate perturbations generally preserve competitive AUC values, indicating that the framework is not driven by an isolated lucky hyperparameter choice. This supports the robustness of the OT-based scoring design under the source-only evaluation protocol.
\subsection{MITRE ATT\&CK Contextualization}
\label{sec:mitre_mapping}

To assess threat relevance, we map representative high-ranked processes to MITRE ATT\&CK\footnote{\texttt{https://attack.mitre.org/}} techniques. This mapping is used only for post-hoc explanation, not for training, scoring, calibration, or fusion. It connects low-level provenance events with adversarial intent and helps SOC analysts interpret why a ranked process is security-relevant. Representative detections align with common attack stages such as \emph{Execution}, \emph{Persistence}, \emph{Privilege Escalation}, \emph{Defense Evasion}, and \emph{Command and Control}. Linux events such as \texttt{event\_open}, \texttt{event\_exec}, \texttt{event\_fork}, and \texttt{event\_rename} suggest suspicious execution or anti-forensics, consistent with T1036.003, T1055, and T1543.003. Windows patterns involving \texttt{event\_regmod}, \texttt{event\_createproc}, disk writes, network connections, or \texttt{event\_schtasks} indicate registry persistence, command execution, scheduled-task persistence, payload staging, or C2 activity, consistent with T1547.001, T1059, T1105, T1053.005, and T1027. BSD permission change-and-execution patterns map to T1222.002 and T1548.001, while Android APK installation, network transmission, and wakelock behavior reflect spyware-like persistence or exfiltration patterns, e.g., T1476, T1409, and T1547. Overall, ATT\&CK contextualization shows that the detector surfaces recognizable adversarial tradecraft rather than arbitrary statistical outliers.
\subsubsection{Lessons Learned:}
Our study identifies and addresses six major challenges in achieving effective cross-OS transfer for APT detection. First, we overcome feature misalignment across OSs by translating raw activity into normalized natural language, enabling semantic abstraction through pretrained LLMs. Second, we leverage the DARPA TC dataset—unique in offering multi-OS APT traces —supporting 12 transfer scenarios. Third, to mitigate the absence of target labels, we rely solely on source supervision and unsupervised scoring (OT, VGAE, similarity) in the target. Fourth, to handle poor generalization of classical models (e.g., AE, IF), we learn source manifolds and project target embeddings via Optimal Transport. Fifth, we fuse heterogeneous signals (semantic, structural, geometric) using a late-stage max aggregation—striking a balance between simplicity, robustness, and deployability. Finally, we reduce GPT translation costs by caching translated process actions, ensuring scalable and efficient cross-domain modeling.
Together, these solutions enable a robust source-only, target-unsupervised transfer framework for APT detection under real-world cross-platform constraints.
\subsection{Limitations}
\label{subsec:limitations}

Although the proposed framework shows strong performance under source-only cross-OS transfer, several limitations should be acknowledged. First, the method assumes access to labeled source-domain data in order to identify source-normal processes and construct the normal reference memory. Therefore, the framework is target-unsupervised but not fully label-free: source labels are still required to define the benign reference distribution. In environments where even source-domain labels are noisy or unavailable, additional mechanisms for robust normal-reference construction would be needed.

Second, the semantic alignment step relies on controlled keyword-level translation to reduce cross-OS vocabulary mismatch. Although the translation prompt is constrained to preserve observed events, IP addresses, ports, process identifiers, and action order, imperfect mappings may still occur when an OS-specific artifact has no clear equivalent in the target platform. We reduce this risk by using translation only as a vocabulary canonicalization step, caching repeated mappings, and retaining tokens when no reliable equivalent exists. Nevertheless, translation quality may influence the embedding geometry and therefore the downstream anomaly scores.

Third, the OT-based score depends on the quality of the embedding space. If semantically related source and target behaviors are not placed near compatible regions of the representation space, the barycentric projection may become less informative. This limitation is partly mitigated by evaluating multiple language-model backbones and by combining OT with structural and semantic evidence, but the framework may still be affected by embedding bias or poor representation of rare system behaviors.

Fourth, the structural channel depends on the construction of the source-normal \(k\)-nearest-neighbor graph. The choice of neighborhood size and graph density can affect VGAE reconstruction behavior, especially when the source-normal memory is small or highly imbalanced. We address this by using a density-controlled graph construction and by evaluating sensitivity to key parameters, but more adaptive graph construction strategies could further improve robustness.

Fifth, while the DARPA Transparent Computing data provides one of the most suitable public benchmarks for cross-OS APT detection, it remains a finite benchmark with a limited number of attack scenarios. Real SOC deployments may involve additional operating systems, evolving software stacks, noisier telemetry, and adversaries whose behaviors differ from those represented in the dataset. Therefore, further validation on additional provenance datasets and operational traces would strengthen external validity.

Finally, the proposed framework improves ranking quality but does not by itself replace analyst judgment. The MITRE ATT\&CK contextualization provides post-hoc interpretability for selected high-ranked processes, but it is not a formal causal attribution mechanism. Future work should integrate richer explanation modules, analyst feedback, and online adaptation mechanisms while preserving the source-only constraint.

These limitations do not undermine the central empirical finding: source-only cross-OS APT detection can be substantially improved by combining semantic abstraction, structural modeling, and OT-based geometric anomaly scoring under a strict zero-target-label protocol.
\section{Conclusion and Future Work}
We introduced a source-only, target-unsupervised framework for cross-OS APT detection, that leverages LLM-generated behavioral descriptions, optimal transport alignment, and multi-path anomaly scoring. Our method generalizes from a single labeled source OS to multiple unseen targets, eliminating the need for costly retraining or domain adaptation. Through extensive evaluation on DARPA TC data across 12 OS pairs, we demonstrated the superiority of our approach over classical transfer baselines. For future work, we plan to explore few-shot adaptation via user feedback, causal modeling of cross-domain process behavior, and integration with real-time security event streams to support continuous learning under evolving attack surfaces.

\section*{Conflicts of Interest}
The authors declare no conflict of interest.
\bibliographystyle{ieeetr}
\bibliography{biblio}

\end{document}